\documentclass[lettersize, journal]{IEEEtran}
\usepackage{times}

\usepackage[numbers]{natbib}
\usepackage{multicol}
\usepackage[bookmarks=true]{hyperref}

\usepackage{amsmath} 
\usepackage{amssymb}  
\usepackage{amsfonts}
\usepackage{mathrsfs}
\usepackage{wrapfig}
\usepackage{graphicx}
\usepackage{textcomp}
\usepackage{xcolor}
\usepackage[capitalise]{cleveref}
\usepackage[list=on,listformat=simple]{subcaption}
\usepackage{preambles}
\usepackage{algorithm} 
\usepackage{algpseudocode}
\usepackage{booktabs}

\graphicspath{{./figs/}}

\newcommand*{\revised}{\textcolor{black}}
\newcommand*{\revisedgc}{\textcolor{black}}

\newcommand*{\colorlink}{\textcolor{violet}}

\begin{document}

\title{Learning Terrain Aware Bipedal Locomotion via Reduced Dimensional Perceptual Representations}

\author{Guillermo A. Castillo*$^{1}$,  Himanshu Lodha*$^{1}$, and Ayonga Hereid$^{2}$
\thanks{This work was supported in part by the National Science Foundation
under grant FRR-21441568.}%
\thanks{$^{*}$Equal contribution.}%
\thanks{$^{1}$Electrical and Computer Engineering, The Ohio State University, Columbus, OH, USA;  {\tt\footnotesize (castillomartinez.2, lodha.11)@osu.edu.}}
\thanks{$^{2}$Mechanical and Aerospace Engineering, The Ohio State University, Columbus, OH, USA. {\tt\footnotesize hereid.1@osu.edu.}}%
}


\IEEEpubid{0000--0000/00\$00.00~\copyright~2025 IEEE}

\maketitle
\begin{abstract}
This work introduces a hierarchical strategy for terrain-aware bipedal locomotion that integrates reduced-dimensional perceptual representations to enhance reinforcement learning (RL)-based high-level (HL) policies for real-time gait generation. Unlike end-to-end approaches, our framework leverages latent terrain encodings via a Convolutional Variational Autoencoder (CNN-VAE) alongside reduced-order robot dynamics, optimizing the locomotion decision process with a compact state. We systematically analyze the impact of latent space dimensionality on learning efficiency and policy robustness. \revisedgc{Additionally, we extend our method to be history-aware, incorporating sequences of recent terrain observations into the latent representation to improve robustness. To address real-world feasibility, we introduce a distillation method to learn the latent representation directly from depth camera images and provide preliminary hardware validation by comparing simulated and real sensor data. We further validate our framework using the high-fidelity Agility Robotics (AR) simulator, incorporating realistic sensor noise, state estimation, and actuator dynamics. The results confirm the robustness and adaptability of our method, underscoring its potential for hardware deployment.}
\end{abstract}

\IEEEpeerreviewmaketitle

\section{Introduction}

One of the main advantages of legged robots over their wheeled counterparts is their potential to navigate challenging and unstructured environments. The early stage of legged locomotion research focused on \textbf{\emph{blind locomotion}} where robots were designed to move without real-time perceptual feedback from their environment. These systems relied heavily on pre-programmed movements and robust locomotion controllers to navigate their surroundings. However, as anyone who has observed the effortless grace of animals and humans to traverse rugged terrains can attest, there is an essential difference between simply moving and moving with awareness of one's environment, known as \textbf{\emph{perceptive locomotion}}.

\begin{figure}[t]
    \vspace{2mm}
    \begin{center} 
    \includegraphics[width=1\linewidth]{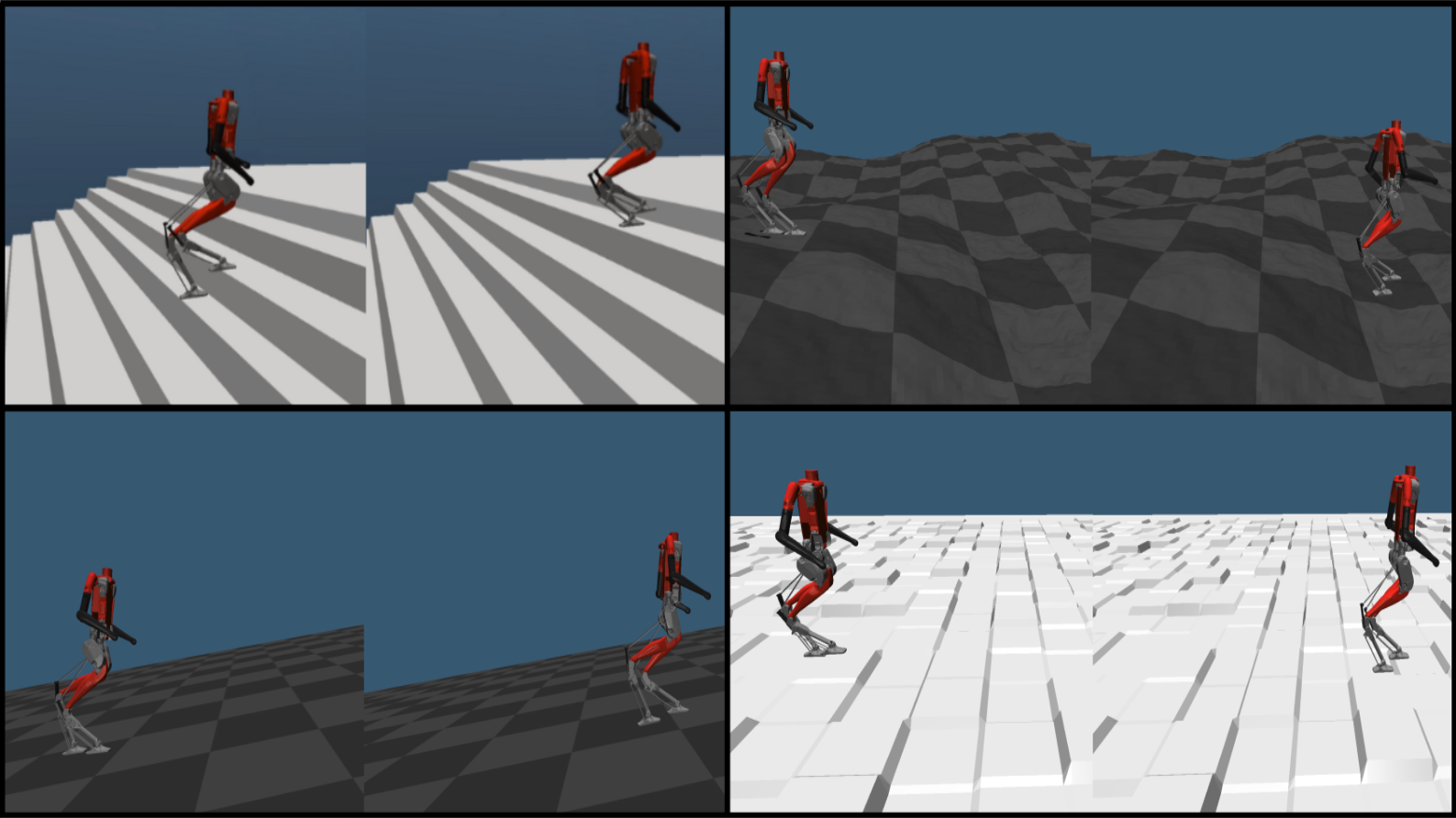}
    \end{center} 
     \vspace{-2mm}
    \caption{Digit walking over challenging terrains (stairs, hills, slopes, and squares) using a terrain-aware locomotion policy.} 
    \label{fig:tileplots_terrain}
\end{figure} 

Most existing work on RL-based controllers for bipedal locomotion has focused on blind locomotion. The impressive robustness of the learned policies allows the robot to walk in challenging terrains such as hills~\cite{xie2018feedback}, slopes~\cite{castillo2022reinforcement}, and even flights of stairs~\cite{siekmann2021blind}.
The shift from blind to perceptive locomotion has enabled robots to see and respond to their environment in real-time. Integrating visual sensors improves stability and safety by allowing adaptive adjustments to gait to account for different terrains and environments. In particular, there has been a growing interest in integrating terrain information as part of the state feedback for training gait policies for legged locomotion, especially on quadruped robots.

\begin{figure*}[ht]
    \vspace{1mm}
    \begin{center}
    \includegraphics[trim={15mm 920mm 15mm 15mm},clip,width=1\linewidth]{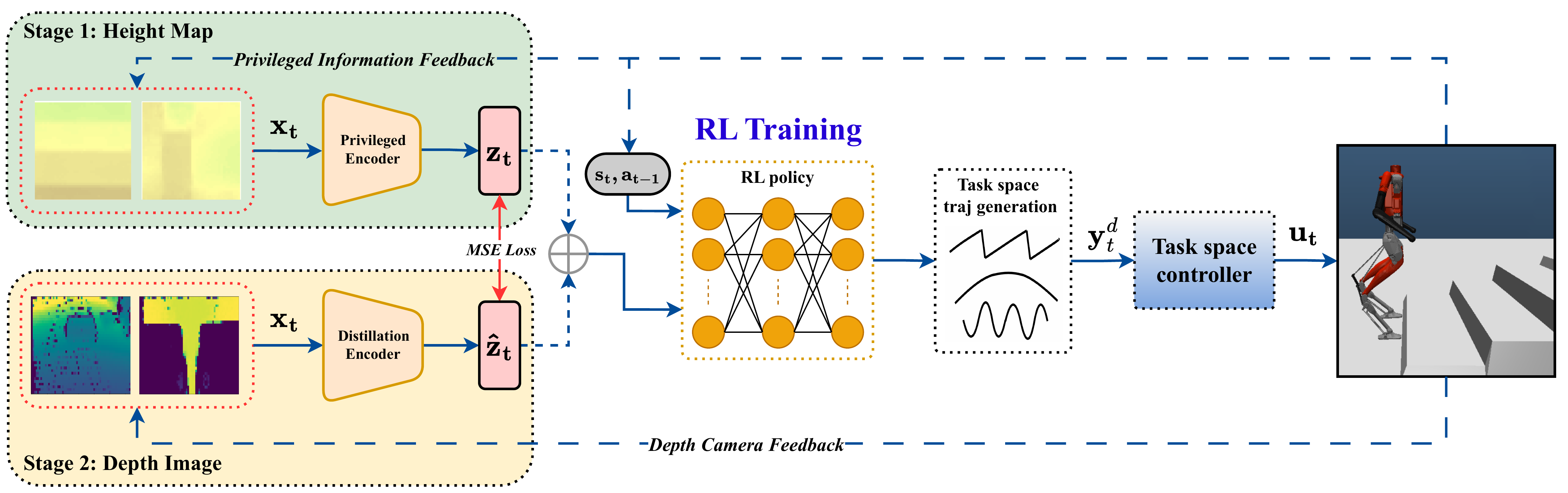}    
    \end{center} 
    \caption{\revisedgc{The hierarchical structure of the proposed framework: a high-level RL policy for gait planning trained with a multi-stage approach, and a low-level controller for trajectory tracking. The privileged encoder uses a CNN-VAE to encode the local height map to a reduced-dimensional latent variable to train terrain-aware perception locomotion policies. In Stage 2, a distillation process replaces the privileged information from the height map with the input from depth cameras by matching the latent representation obtained from these two exteroceptive sources.}}
    \label{fig:overall_framework}
\end{figure*} 

Some of the first attempts to integrate terrain information in an RL policy for bipedal locomotion were implemented in simulations of physics-based animations. In~\cite{peng2015dynamic}, a reduced character state and reduced terrain state were used to train a policy to navigate terrains with steps and gaps in a simulation of a 2D environment. This approach was extended~\cite{peng2016terrain} by using the full height field map and character state in an end-to-end RL framework with a mixture of actor-critic experts updated through temporal difference learning. 
\IEEEpubidadjcol
An extension to the 3D case in simulation was proposed by~\cite{peng2017deeploco}, where Hierarchical Deep Reinforcement Learning was used to train a high-level policy that makes step target decisions based on high-dimensional inputs, including terrain maps or other suitable representations of the surroundings, and a low-level policy that learns to achieve robust walking gaits. 

The works in ~\cite{xie2020allsteps,singh2022learning} address the challenge of walking on irregular terrains using pre-planned footsteps obtained from the environment height field. The RL policy then uses the foothold sequence and the robot's state to compute the joint target positions. A more effective terrain representation is presented in \cite{acero2022learning} using a sparse exteroceptive observation from raycasts along the vertical axis. This approach is efficient, but limited in the number of features it can capture.

\revisedgc{Marum et al.~\cite{vanmarum2023learning} build upon the work in latent terrain representation for quadruped locomotion~\cite{miki2022learning} to train an end-to-end policy to navigate a wide variety of terrains using noisy exteroception. Duan et al.~\cite{duan2023learning} presented a vision-based RL framework for bipedal locomotion, showcasing robust locomotion over challenging terrains with the robot Cassie. A height map expressed in the robot's local frame is used to train an end-to-end RL locomotion policy for stairs and steps of different heights. The height map used to train in simulation is replaced by a height map predictor obtained from depth camera images and the robot state. Gadde et al.~\cite{gadde2025learning} build upon ~\cite{duan2023learning} to replace the height map estimator with a perception encoder trained with a teacher-student approach. The work in \cite{gu2024advancing} and \cite{wang2024combining} present an alternative to visual perception by integrating the terrain information only as privileged information and using the history of the observation data and an Asymmetric Actor-Critic architecture \cite{gu2024advancing} or a student-teacher approach \cite{wang2024combining} to replace the visual perception by encoded latent space that captures the terrain conditions along with the robot dynamics.}

\revisedgc{While these recent advancements push the boundaries of perceptive locomotion through sophisticated attention mechanisms that dynamically select terrain features~\cite{he2025attention}, novel hybrid training paradigms~\cite{zhang2025distillationppo}, and advanced sensor fusion with internal models~\cite{long2025learning}, our work provides a distinct and complementary contribution. These state-of-the-art methods focus on building complex integrated systems, often leveraging rich perception sources like LiDAR-based elevation maps  \cite{long2025learning} or innovating on the training algorithm itself, \cite{zhang2025distillationppo}.}

\revisedgc{Our work, in contrast, addresses a more fundamental and generalizable question: What is the principle of minimal sufficiency for perceptual information in locomotion? Our primary contribution is the development of a new policy architecture and the systematic analysis of the information bottleneck between perception and action. We extend our analysis to history-aware autoencoders, which provide further evidence for our central hypothesis. Moreover, we introduce a distillation process that enables the policy to learn the same compact representation directly from multiple depth camera images—a critical step towards hardware deployment. We leverage a lightweight, hierarchical framework and a simple CNN-VAE not to build the most complex system, but as a precise tool to investigate the trade-off between the dimensionality of the latent space and the resulting policy's performance.}

\revisedgc{This paper extends our prior work, which first established a sample-efficient hierarchical framework for bipedal locomotion~\cite{castillo2023template} and subsequently validated its real-world viability with successful zero-shot sim-to-real transfer on the Digit robot~\cite{weng2023standardized}. While this proprioceptive-based controller proved robust to external disturbances, it was fundamentally "blind" and thus incapable of navigating unstructured terrain.}

\revisedgc{The primary contribution of this work is therefore the integration of a perception module into this proven hierarchical framework. Leveraging insights from our previous research on data-driven latent spaces~\cite{castillo2023datadriven}, we introduce a learned, low-dimensional terrain representation that allows the policy to make informed, terrain-aware decisions. This addition bridges the critical gap from blind disturbance rejection to agile, perceptive locomotion over complex terrains.}


\revisedgc{Therefore, in this work, we propose a perceptive bipedal locomotion framework that combines the versatility of RL-based policies for high-level commands with the robustness of a low-level task space controller and the effectiveness of an efficient latent representation of the terrain height map. One of the key contributions of our work is the systematic analysis of the impact of the dimension of the latent representation of the height map on the efficiency of the learning process, showing that a too small or too large dimension of the latent representation hurts sample efficiency.  This analysis is further extended to history-aware perception, and we introduce a distillation method to learn the same compact representation directly from depth camera images, a critical step for real-world deployment. By focusing on the efficiency of the representation itself, our work provides concrete, empirical evidence that an optimal level of compression exists.} 

The remainder of the paper is organized as follows: Section~\ref{section:terrainlatent} explains the supervised learning approach to encode a latent terrain representation with CNN-VAEs and discusses the importance of the latent space dimension. Section~\ref{sec:method} introduces a hierarchical RL framework for terrain-aware locomotion. \revisedgc{Section~\ref{sec:sim_results} shows simulation results of the proposed framework with ablation studies and baseline comparisons, addressing real-world feasibility through a distillation process with depth images, realistic sensor noise, and additional tests with the high-fidelity Agility Robotics (AR) simulator.}
Finally, Section~\ref{sec:conclusion} briefly concludes the paper and discusses the future directions of our work.

\section{Terrain Representation in Latent Space} \label{section:terrainlatent}
In this section, we introduce our proposed method to learn an adequate representation of the terrain information around the robot that successfully captures the critical features of the ground. The goal is to design a robust locomotion policy, introduced in Section~\ref{sec:method}, that allows humanoid robots to navigate challenging terrains actively. 

\subsection{Terrain data collection} \label{subsec:datacollection}
We use a local height map corresponding to the area of $2 \mathrm{m}^2$ in front of the robot to perceive the terrain around the robot. When using a resolution of $5 \mathrm{cm}^2$, the local terrain height map, $\mathbf{x}$, is represented by a matrix of size $20\times40$, resulting in a total of 800 elements. 

\begin{figure}[t]
    \vspace{2mm}
    \begin{center} 
    \includegraphics[width=1\linewidth]{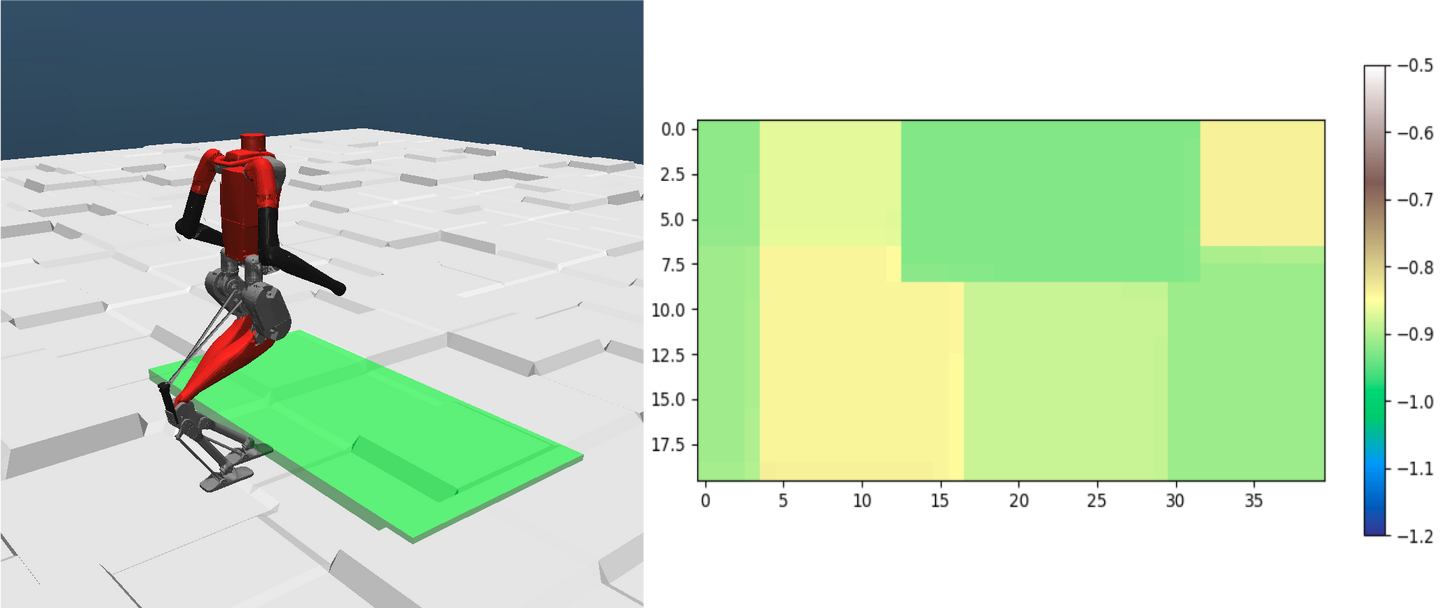}
    \end{center} 
    \caption{World view of the local height map used to detect the terrain around the robot (left) with the corresponding local height map matrix (right) relative to the robot's base. The height map covers an area of $2 m \times 1 m$ at a 5 cm resolution.}
    \label{fig:local_height_map}
\end{figure}

The height map grid resolution choice is based on the width of the Digit robot's feet, which is about 5 cm. It is also aligned with relevant works in literature, where a grid resolution of 5 to 6.5 cm is used to capture features of different terrains effectively \cite{hoeller2023anymal, duan2023learning, vanmarum2023learning}. We show in \figref{fig:local_height_map} a sample of this local height map obtained from a simulation. Using all the elements of the local terrain height map matrix as input for the locomotion policy would result in a significantly large neural network and, consequently, more parameters and larger inference time, even though many of the elements of the height map may not have useful information for the policy to produce effective gait actions.

To train a CNN-VAE of the terrain map, we use a customized simulation environment in MuJoCo~\cite{todorov2012mujoco} to create different terrain profiles, including sloped planes, hills, squared steps, and stairs with various configurations (up, down) and dimensions (width, depth, and height). We collect $60,000$ samples of local height maps for each terrain type.
To collect a diverse dataset of terrain height maps without the existence of a capable locomotion policy, we only simulate the kinematic motion of the robot around the terrain by updating the position of the robot's base in the simulation environment according to randomly sampled velocity while keeping the base height at a height that follows a normal distribution with a mean $0.92$ m and standard deviation $0.1$ m. Thus, the complete dataset $\mathbf{X}$ consists of $360,000$ samples of local terrain height maps $\mathbf{x}$, i.e., $\mathbf{X}=\{\mathbf{x}^{(i)} | i\in[1,360,000]\}$. 


\subsection{Convolutional Variational Autoencoder}
One of CNN-VAEs' primary advantages is their proficiency in handling high-dimensional data, such as large maps, and effectively compressing them into lower-dimensional representations that capture the essential features and structures of the original data through a conditioned probability distribution. 

\begin{figure}[t]
    \vspace{1mm}
    \begin{center}   
    \includegraphics[trim={8mm 885mm 6mm 6mm},clip,width=1\linewidth]{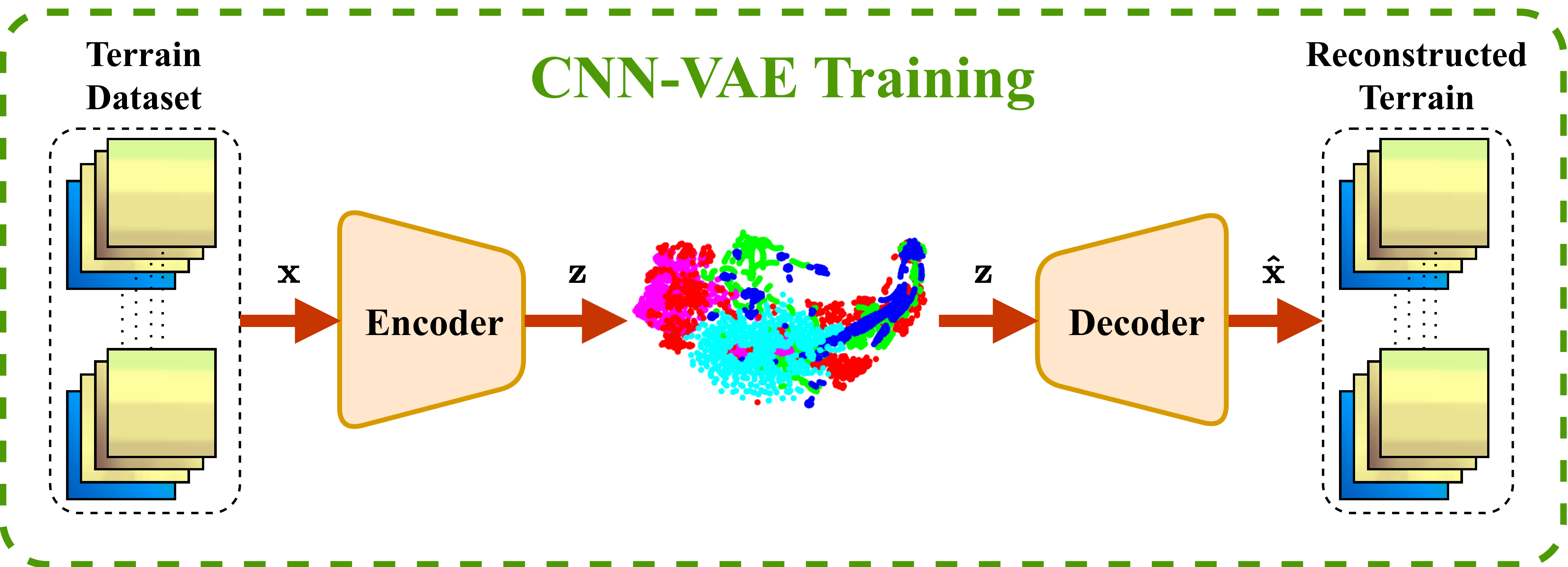}    
    \end{center} 
    \caption{\revisedgc{The CNN-VAE encodes the local height map to a reduced-dimensional latent variable, $\mathbf{z}$, used to train terrain-aware perception locomotion policies with the framework presented in \figref{fig:overall_framework}.}}
    \label{fig:cvae_training}
\end{figure}

In particular, as shown in \figref{fig:cvae_training}, we use a CNN-VAE to encode the terrain height map into a reduced-dimensional latent variable $\mathbf{z} \in \mathbb{R}^m$ to reduce the dimension of terrain information used for locomotion.
In this work, the \textit{encoder} part of the CNN-VAE comprises three convolutional layers followed by two fully connected layers. The convolutional layers progressively reduce the spatial dimensions of the input while increasing the depth of the feature maps, with 32, 64, and 128 channels, respectively, each using a kernel size of 4, a stride of 2, and a padding of 1.
After the convolutional layers, the output is flattened and passed through two fully connected layers. Specifically, these layers output the mean $\mu$ and variance $\sigma$ of the prior distribution, both sized according to the predefined latent variable dimension $m$. Then, the latent random variable $\mathbf{z}$ can be expressed as a deterministic variable
\begin{align} \label{eq:reparameterization}
    \mathbf{z}=g_{\theta}(\epsilon, \mathbf{x}),
\end{align}
\revisedgc{where $\mathbf{x}$ is the sample vector corresponding to the local height map, $\theta$ represents the learnable parameters (weights and biases) of the encoder network, $g_\theta(\cdot)$ is the encoder function parameterized by $\theta$, and $\epsilon$ is an auxiliary random variable with an independent marginal probability distribution.} \revisedgc{In this context, the term latent random variable refers to the hidden variables (the elements of the vector z) that are not directly observed but are instead inferred from the input data (the terrain height map). In a VAE, these variables are treated probabilistically, allowing the model to capture a distribution of the underlying terrain features.} Therefore, if we choose $\epsilon$ to be the univariate Gaussian distribution $\mathcal{N}(0,1)$, the latent random variable $\mathbf{z}$ is determined by 
\begin{align} \label{eq:reparameterization_explicit}
    \mathbf{z}= \mathbf{\mu} + \mathbf{\sigma} \mathbf{\epsilon}.
\end{align}

This method, known as the reparameterization trick, allows backpropagation through random sampling processes, which is essential to train VAEs through standard stochastic gradient descent methods. 
The encoder part of the CNN-VAE efficiently reduces the dimensionality of the input data and captures its essential features in a form conducive to generative tasks. By learning this compact latent representation, the CNN-VAE can effectively generate a probability distribution from which one can reconstruct the local height map samples and even generate new, unseen local height maps that share statistical properties with the training data.

\subsection{Reconstruction of the height map}
The \textit{decoder} is the closed-form parameterized function that maps from the latent space back to the full-order state. This function is defined by
\begin{align} \label{eq:decoder}
    \hat{\mathbf{x}} = d_{\phi}(\mathbf{z}),
\end{align}
\revisedgc{where $\phi$ represents the parameters of the decoder network, $d_{\phi}$ is the decoder function parameterized by $\phi$, $\mathbf{z}$ is the encoded latent variable, and $\hat{\mathbf{x}}$ is the reconstruction of the original input data $\mathbf{x}$.}

\begin{figure}[t]
    \begin{center} 
    \includegraphics[width=0.9\linewidth]{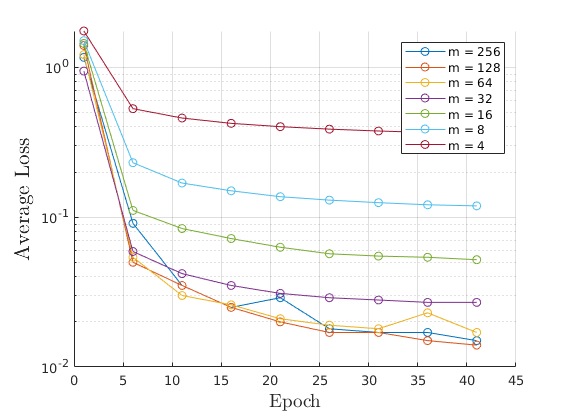}
    \end{center} 
    \caption{Latent representation learned by the CNN-VAE for different dimensions of the latent variable $m$. Larger latent sizes (e.g., 64, 128) converge faster and to a lower loss, indicating better reconstruction, showing diminishing returns for $m\geq32$, whereas minimal latent dimensions (e.g., $m=4$) show significant error. This suggests that overly compressing the terrain representation hurts accuracy, while moderately sized latents effectively balance compactness and fidelity.} 
    \label{fig:loss_training}
\end{figure}

The CNN-VAE is trained by minimizing the standard $\beta$-VAE loss $\mathcal{L}$, which consists of reconstruction loss and the Kullback-Leibler (KL) divergence 
as the latent loss~\cite{kingma2014autoencoding}. Then, the VAE loss is formulated as 
\begin{align}\label{eq:VAE_loss}
    \mathcal{L} = \text{MSE}(\mathbf{x}^{(i)}, \hat{\mathbf{x}}^{(i)}) + \beta D_{\text{KL}}(q(\mathbf{z}^{(i)}|\mathbf{x}^{(i)})||p(\mathbf{z}^{(i)}))
\end{align}
where $\hat{\mathbf{x}}^{(i)}$ is the reconstructed height map, $p(\mathbf{z}^{(i)})$ is the prior distribution parameterized by the Gaussian distribution $\epsilon$, and  $q(\mathbf{z}^{(i)}|\mathbf{x}^{(i)})$ is the posterior distribution of the latent variable $\mathbf{z^{(i)}}$ given $\mathbf{x}^{(i)}$.
The autoencoder is trained using Adam optimizer \cite{kingma2015adam} with a learning rate of $0.001$ and a batch size $B$ of $256$. The autoencoder is trained for 40 epochs in a 12-core CPU machine with an NVIDIA RTX 2080 GPU.

There is no rule of thumb for the proper size of the latent state used to capture the encoder input's features fully. On the one hand, a latent variable of large dimension allows a better reconstruction of the local height map. Conversely, a smaller dimension of the latent variable enables more efficient encoding, resulting in more compact networks for the VAE and the locomotion policy. To analyze the trade-off between these two properties, we conduct an ablation study with different values of the latent space dimension. In ~\figref{fig:loss_training}, we show the loss $\mathcal{L}$ during training of the CNN-VAE for different values of $m$. The latent dimensions 256, 128, and 64 show the most rapid decrease in loss, indicating efficient learning and improved ability to capture data features. The latent dimensions 64 and 32 clearly balance model complexity and learning efficiency. The smaller dimensions 8 and 4 exhibit a higher average loss, which means the model is not complex enough to capture all the necessary data features. 
 
\begin{figure}[t]
    \begin{center} 
    \includegraphics[trim={35mm 10mm 45mm 10mm},clip,width=1\linewidth]{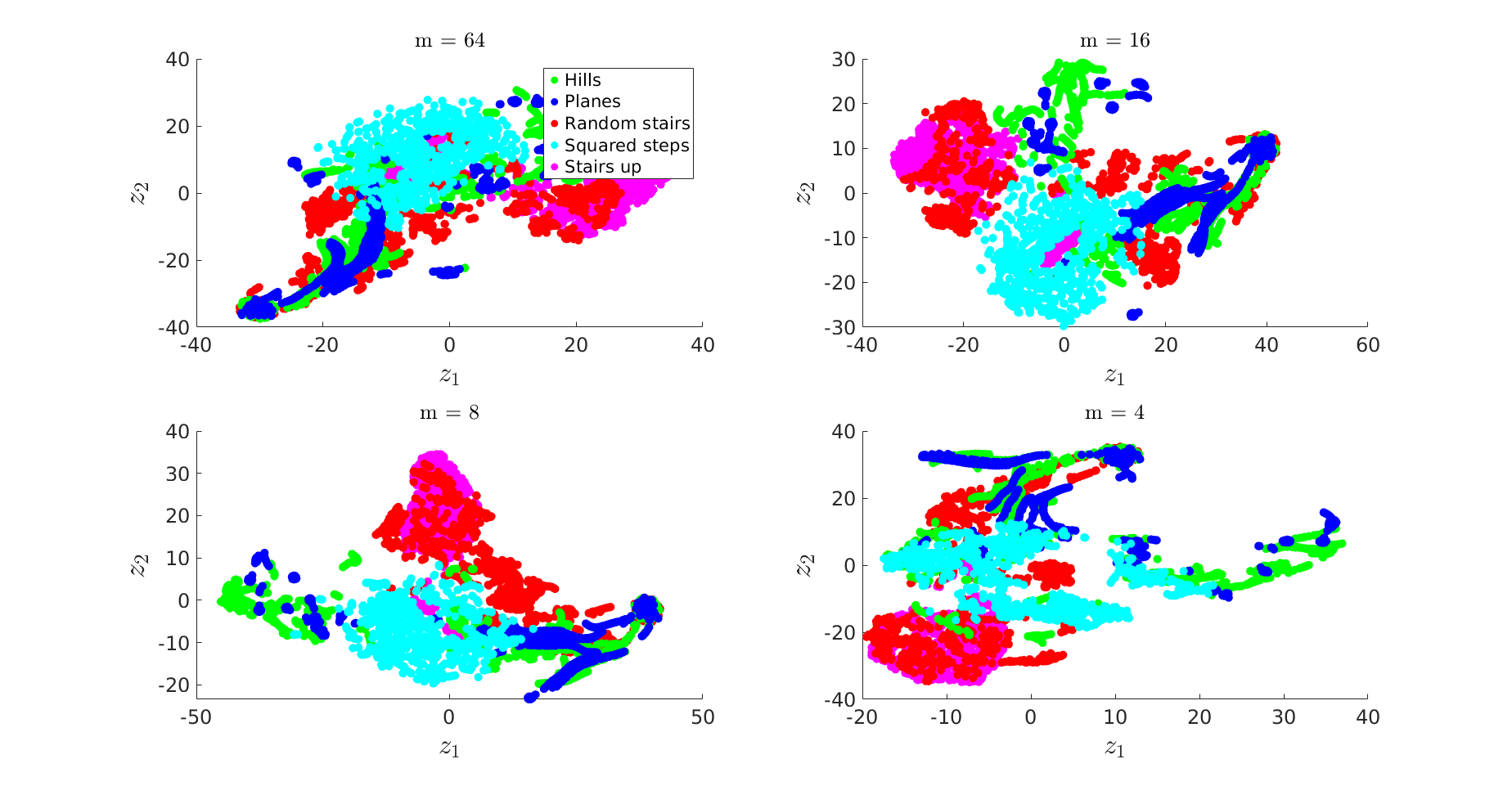}
    \end{center} 
    \caption{t-SNE of the latent representation learned by the CNN-VAE for different dimensions of the latent variable $\mathbf{z}$.
    t-SNE visualization of the learned latent space for different latent dimensions m. For sufficiently large latent dimensions $(m\geq16)$, the latent vectors form well-separated clusters corresponding to distinct terrain types, indicating that the VAE has retained meaningful terrain features. In contrast, with a minimal latent ($m=4$), the clusters – particularly for complex terrains like “squared steps” – are poorly formed, showing that important details are lost when the latent space is too limited.
    }
    \label{fig:latent_space_comparison}
\end{figure}

In addition, we apply t-Distributed Stochastic Neighbor Embedding (t-SNE) to the latent representations of different dimensions to analyze the structure of the encoded data. These results, presented in ~\figref{fig:latent_space_comparison}, demonstrate that for $m \geq 16$, the latent space retains a consistent and meaningful structure, with clusters representing distinct terrain types. However, the structure deteriorates significantly for $m = 4$, particularly for the squared steps terrain. This terrain exhibits the highest degree of irregularity and complexity, is the most challenging to encode, and requires more features to capture its structure accurately. This behavior is expected because more irregular terrains demand a higher-dimensional latent space to retain their essential characteristics.
This analysis highlights that a latent dimension as low as $m =8$ can still effectively capture the key features of the terrain height map, achieving a balance between compactness and representational power. Consequently, we choose $m = 16$ as a good trade-off between feature capturability and reconstruction error for the CNN-VAE during the training of our locomotion policy, described in the next section.

\revisedgc{\subsection{History-Aware Perception} \label{subsec:history_method}
Recent advancements in perceptive history for legged locomotion demonstrate that integrating historical data enhances robots' adaptability to changing environments, optimizing movement, and improving navigation efficiency. 
However, seamlessly combining historical and real-time data across multiple modalities could significantly increase the input dimension of end-to-end RL approaches. Therefore, analyzing the optimality and efficiency of the latent representation is even more relevant as it could significantly impact the design and efficiency of the framework. Thus, in this section, we explore the use of reduced-order representations developed in Section \ref{section:terrainlatent} to capture the features of the terrain along a history of local height maps.}

\revisedgc{The CNN-VAE architecture illustrated in \figref{fig:cvae_training} can be easily modified to integrate the history of the local height maps. While its fundamental structure remains unchanged,
adjustments are made to accommodate $n$-inputs and $n$-outputs, corresponding to the last $n$ history steps, enabling an $n$-to-$n$ mapping. Since the local height map is a single-channel image, each of the $n$ inputs is stacked in a channel to obtain an $n$-channel image, with the output following the same structure.}

\revisedgc{We experiment with two modified structures of the framework, $n$-to-$n$ and $n$-to-$1$. In the $n$-to-$n$ case, the latent representation captures the terrain features associated with the latest $n$ terrain height maps and reconstructs the same $n$ height maps from the latent variable. In the $n$-to-$1$ case, the latent variable is used to reconstruct only the single latest height map of the sequence in the $n$ inputs.  Similar to what was observed in Section II.C, there is a tradeoff in the dimension of the latent representation and the reconstruction accuracy, where increasing the size of the latent representation does not improve the reconstruction loss. }

\begin{figure}[t]
    \begin{center} 
    \includegraphics[trim={0mm, 0mm, 0mm, 0mm},clip,width=0.9\columnwidth]{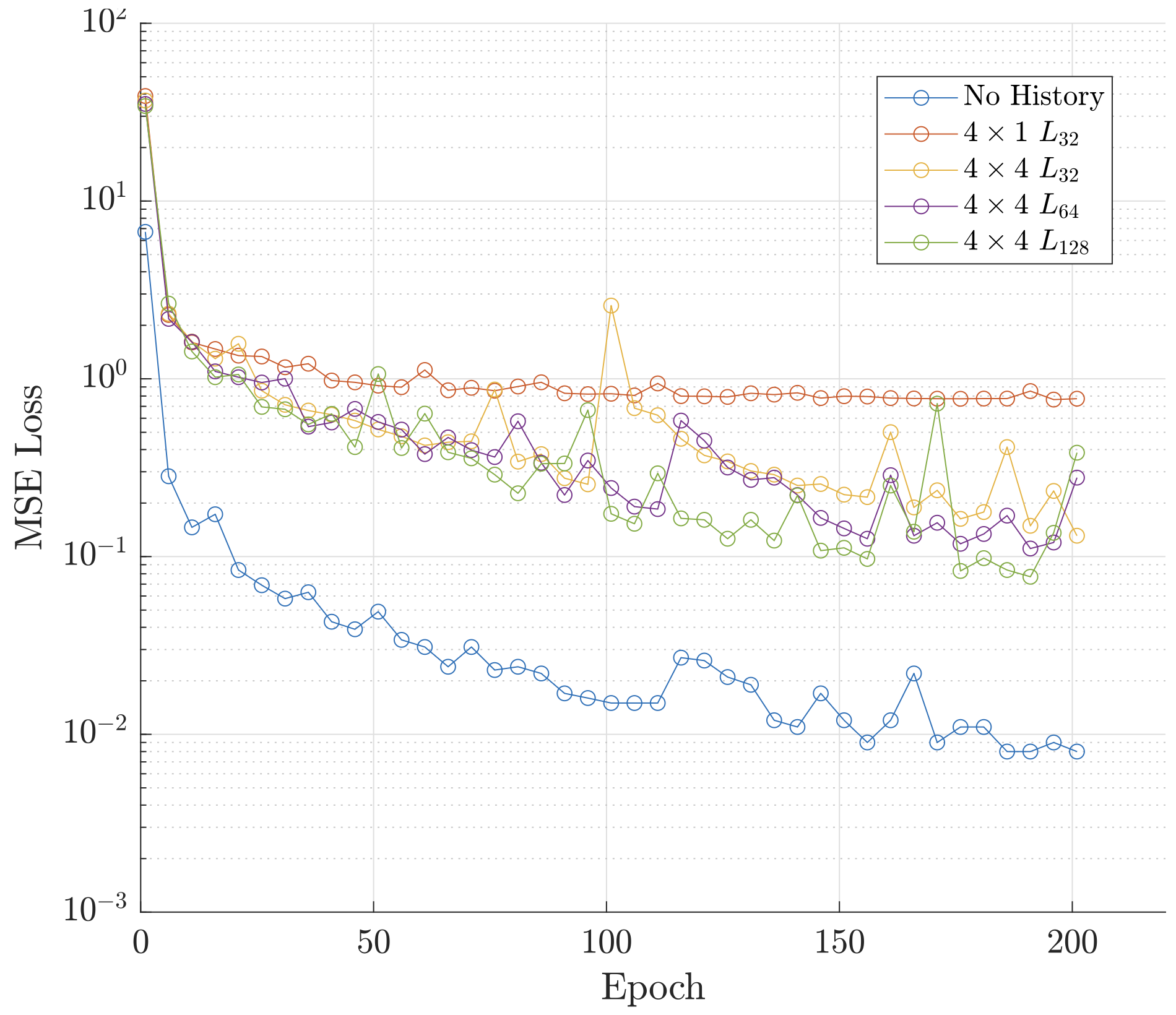}
    \end{center} 
    \caption{\revisedgc{Training loss curves for CNN-VAE models incorporating terrain history and different latent sizes. The same trade-off emerges: increasing the latent dimension beyond a certain point yields diminishing improvements.}}
    \label{fig:CVAE_history_train_loss_compare}
\end{figure}

\revised{Section \ref{subsec:history_results} presents an ablation study that explores the impact of the dimension of the latent representation of the height map with and without the history of the terrain height map.}

\section{Hierarchical terrain-aware bipedal locomotion}\label{sec:method}


Building on the success of reduced-order models for online generation of HL trajectories \cite{gong2021onestep, castillo2023template}, we employ reinforcement learning to train a high-level planner policy that harnesses an effective representation of the robot's dynamics and the terrain information. As shown in Fig.~\ref{fig:overall_framework}, the proposed high-level RL policy takes as input a latent space encoding learned from the local height map of the terrain together with a reduced-order representation of the robot's states inspired by the Linear Inverted Pendulum (LIP) model and the state of the swing foot of the robot. The output of the RL policy is a set of task space commands used to generate online task space trajectories for the robot's base and end-effectors. The low-level controller is a model-based whole-body controller used to guarantee the tracking performance of the desired task-space trajectories. \revisedgc{The proposed hierarchical framework allows for replacing the latent variable encoded from the terrain with a latent variable encoded from depth images through an additional distillation stage. The details of the distillation process for the latent space reconstruction from depth images are presented in Section \ref{subsec:distillation}.}


\subsection{Reinforcement Learning for High-Level Planning}
The problem of determining a motion policy for bipedal robots can be modeled as a Markov Decision Process (MDP). The stochastic transition of the MDP process captures the random sampling of initial states in the policy training and dynamics uncertainty due to model mismatch and random interactions with the environment (e.g., early ground impacts).

\subsection{Reduced-Order State Space} 
In this work, we leverage the insights provided by template models to regulate the walking speed of biped robots~\cite{gong2021onestep}. Inspired by the success of~\cite{castillo2022reinforcement, castillo2023template} in using template-based models, we select the state 
\begin{align}
    \mathbf{s} = (\mathbf{x}_\text{b}, h_\text{b}, \mathbf{e}_{\bar{v}}, v_x^d, v_y^d, \mathbf{p_{sw}}, \mathbf{v_{sw}}, \mathbf{z}, \mathbf{a}_{t-1}),
\end{align}
where $\mathbf{x}_\text{b} = (x, y, \dot{x}, \dot{y})$ is the LIP state composed of the robot's base position relative to the stance foot and the base velocity, $h_\text{b}$ is the robot's base height relative to the stance foot, $\mathbf{e}_{\bar{v}}=(e_{\bar{v}_x},e_{\bar{v}_y})$ is the error between the average velocity of the robot's base $(\bar{v}_x, \bar{v}_y)$ and the commanded robot's velocity $(v_x^d, v_y^d)$, $\mathbf{p_{sw}}, \mathbf{v_{sw}}$ are the 3D position and velocity of the robot's swing foot, $\mathbf{z}$ is the latent variable encoded from the local height map centered at the robot's base introduced in Section ~\ref{section:terrainlatent}., and $\mathbf{a}_{t-1}$ is the last policy action. \revisedgc{The positions and velocities of the robot's base and swing foot are expressed in the frame coordinates of the robot's stance foot. All variables correspond to the data at time $t$ unless explicitly denoted with the subscript $t$ as in the last policy action $\mathbf{a}_{t-1}$.}


\begin{figure}[t]
    \vspace{2mm} 
    \begin{center}
    \includegraphics[width=0.9\linewidth]{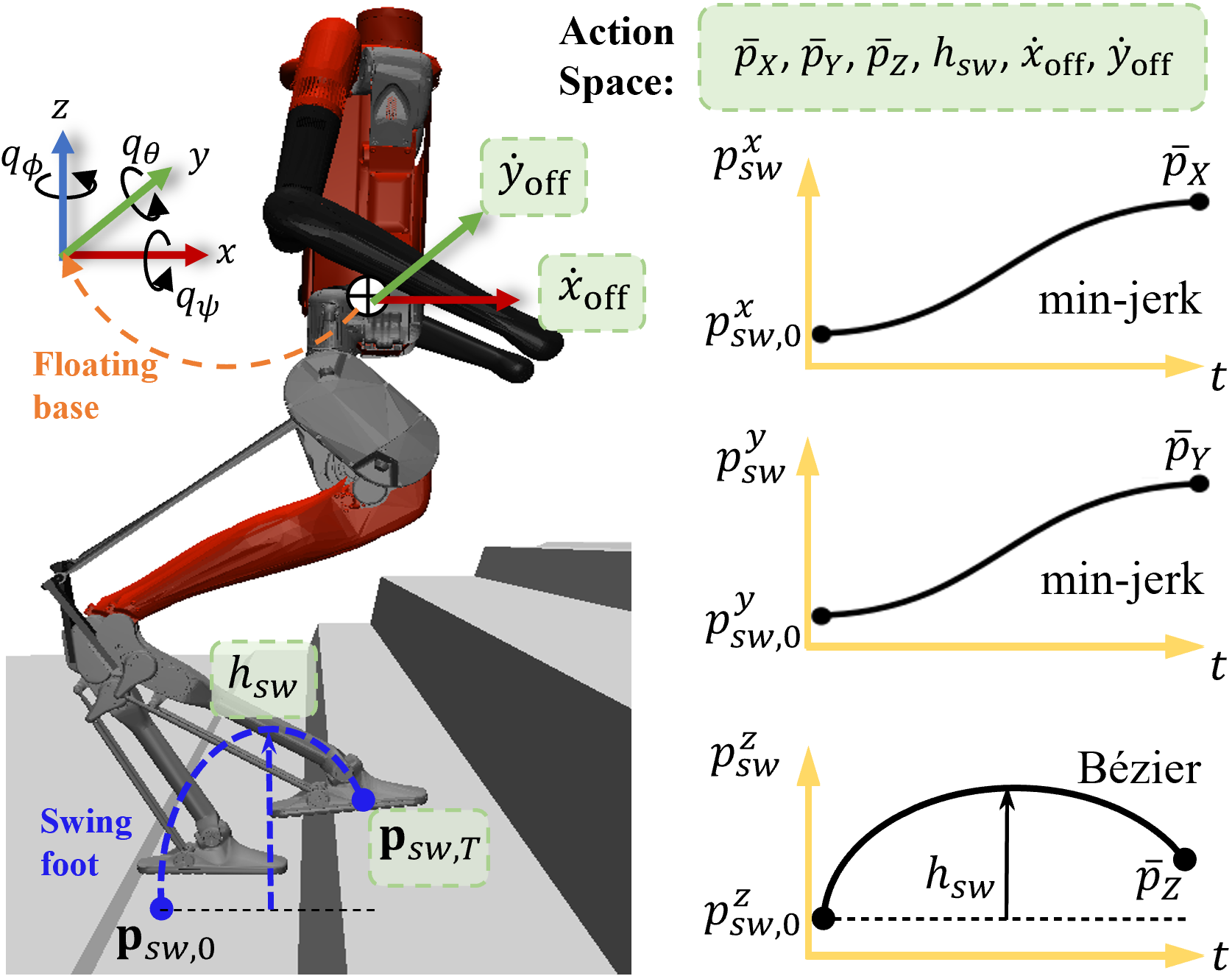}
    \end{center} 
    \caption{Action space representation for the RL policy. The policy outputs target trajectories in task space for swing foot position and base velocity offset, which the low-level controller then executes.} 
    \label{fig:action_space}
    \vspace{-2mm}
\end{figure} 

\subsection{Task Space Actions} \label{subsec:task_space_actions}
The action $\mathbf{a} \in \mathcal{A}$ is chosen to be 
\begin{align}
    \mathbf{a} = (\bar{p}_{X}, \bar{p}_{Y}, \bar{p}_{Z}, h_{\mathrm{sw}}, \dot{x}_{\mathrm{off}}, \dot{y}_{\mathrm{off}}) 
\end{align}
where $\bar{\mathbf{p}} = [\bar{p}_{X},\bar{p}_{Y},\bar{p}_{Z}]^T$ correspond to the landing position of the swing foot w.r.t. the robot's base at the end of the swing phase $T$, $h_{\text{sw}}$ is swing foot clearance, and $\dot{x}_{\text{off}}, \dot{y}_{\text{off}}$ are an offset added to the commanded speed of the robot. This selection of the action space encourages the policy's flexibility to exploit the bipedal robot's natural nonlinear dynamics and enhance the policy's robustness under challenging terrains, disturbances, and sudden speed changes caused by irregularities in the terrain, i.e., tripping over. \figref{fig:action_space} illustrates the selection of the action space on the robot Digit. \revisedgc{The action space of the policy is updated at 33 Hz.} 

Given the desired set of policy actions, the trajectory generation module transforms the policy action $a$ into smooth task-space trajectories for the robot's base and end-effectors. Specifically, as shown in \figref{fig:action_space}, at a time $t \in [0, T]$ the trajectories for the swing foot $p^x_{\text{sw}}(t,\mathbf{a})$ and $p^y_{\text{sw}}(t,\mathbf{a})$ are generated using a minimum jerk trajectory connecting initial foot positions with target foot positions from the policy action. $p^z_{\text{sw}}(t,\mathbf{a})$ is generated using a Bézier polynomial with five control points, with its maximum value corresponding to the height of the swing foot $h_{\text{sw}}$. The initial foot positions are computed at every touchdown event and kept constant throughout the step.

\subsection{Low-level task space controller}
The desired task space trajectories derived from the policy outputs are tracked using task-space inverse dynamics (TSID) with a quadratic programming formulation. We follow the TSID formulation in \cite{yang2022improved}, which considers the constrained dynamics of closed kinematic chains such as the ones in Digit's legs. Here, we only present the problem formulation and refer the interested reader to ~\cite{yang2022improved} for more details.

Consider a bipedal robot with configuration space $\mathcal{Q}\subset\mathbb{R}^n$ and generalized coordinates $q \in \mathcal{Q}$, the equations of motion of the constrained dynamics are given by:
\begin{align}
    \label{eq:full-order-model}
    M(q)\ddot{q} + H(q,\dot{q}) &= B\tau + J_c^T(q) f_c  + N(q)\lambda, \\
    N(q)\ddot{q} + \dot{N}(q,\dot{q})\dot{q} &= 0,  \\
    J_c(q)\ddot{q} + \dot{J}(q,\dot{q})\dot{q} &= 0,  
    \label{eq:holonomic-constraints}
\end{align} 
where $M(q)$ is the inertia matrix, $H(q,\dot{q}) = C(q,\dot{q})\dot{q} + G(q) + F$ is the vector sum of the Coriolis, centripetal, gravitational, and additional non-conservative forces, $B$ is the actuation matrix, $\tau \in \mathbb{R}^{m}$ is the torque inputs at the actuated joints, $J_c(q)$ is the contact Jacobian, $f_c \in \mathbb{R}^{3n_c}$ collects all external contact forces with $n_c$ denoting the number of contacts. Moreover, $N(q) = J_1(q) - J_2(q)$ is the constraint Jacobian matrix, and $\lambda$ is the constrained force due to the closed kinematic chain.

Given the current state \((q, \dot{q})\) of the robot and its task-space references \((X_i^*, \mathcal{V}_i^*, \mathcal{A}_i^*, f_c^*)\), the TSID for the system with constrained dynamics 
is formulated as a quadratic programming problem:

\begin{equation}
\min_{\ddot{q}, f_c, \lambda} \sum_{i} \| J_i \ddot{q} + \dot{J}_i \dot{q} - \mathcal{A}_i \|_{Q_i} + \| f_c - f_c^* \|_{R_f} + \| \lambda \|_{R_\lambda}
\end{equation}
\vspace{-2mm}
\begin{align*}
    \mathrm{st.}\quad   J_c(q) \ddot{q} + \dot{J}_c(q) \dot{q} &= 0,  \tag{contact constraints} \\
    N(q) \ddot{q} + \dot{N}(q) \dot{q} &= 0,  \tag{loop-closure constraints} \\
    f_c &\in \mathcal{F},  \tag{friction cone}\\
    \tau(\ddot{q}, f_c, \lambda) &\in \mathcal{T},  \tag{torque limits} 
\end{align*}
where the weighting matrices \((Q_i, R_f, R_\lambda)\) are positive definite, 
$\tau \in \mathbb{R}^{m}$ is the torque computed by the robot dynamics given $(\ddot{q}, f_c, \lambda)$, $J_i$ is the geometric jacobian of task space references, and $\mathcal{A}_i$ represents the desired spatial acceleration with state feedback, and it is defined by:
\[
\mathcal{A}_i = \mathcal{A}_i^* + K_p \log(X_{m,i}^T X_i^*) + K_d (\mathcal{V}_i^* - \mathcal{V}_{m,i}), 
\]
where $X_{m,i}$ and $\mathcal{V}_{m,i}$ correspond to each task's measured pose and spatial velocity. \(K_p\), \(K_d\) could be seen as the stiffness and damping of a system, and both are positive definite matrices.

The task space references are determined by the desired pose $X_i^*$, spatial velocity $\mathcal{V}_i^*$, and spatial acceleration $\mathcal{A}_i^*$ of the left foot, right foot, and floating base of the robot. The position and linear velocity for the swing foot are obtained from the trajectories generated from the parameters in the action space of the RL policy as discussed in \secref{subsec:task_space_actions}, e.g., $(\mathbf{p}_{\text{sw}}(t,\mathbf{a}), \dot{\mathbf{p}}_{\text{sw}}(t,\mathbf{a}), \ddot{\mathbf{p}}_{\text{sw}}(t,\mathbf{a}) )$. The robot's base velocity is also obtained from the policy actions $(\dot{x}_{\mathrm{off}}, \dot{y}_{\mathrm{off}})$. However, the base's position is not being tracked to reduce disturbances caused by position errors due to discontinuities in the terrain. The roll and pitch orientation and angular velocities of the task space references are zero, while the heading angle determines the yaw orientation. The stance foot target is equal to its current position.
Finally, the force task reference $f_c^*$ is computed using the centroidal dynamics following the approach in \cite{wensing2016improved}.
\subsection{Rewards}
The reward function adopted in this work is designed to exploit the privileged information from the terrain's height map to shape the motion of the robot's swing foot while keeping track of the desired walking speed and reducing the variation of the policy actions between each iteration. More specifically, we define the reward function
\begin{align}
    \label{eq:reward}
    \mathbf{r} = \mathbf{w}^T [ r_{v_x}, r_{v_y},r_{sw_x},r_{sw_y},r_{sw_z},r_{sw_h},r_{sw_f}, r_{a}]^T,
\end{align}
with 
\vspace{-3mm}
\begin{align}
    \label{eq:reward_detailed}
    r_{a} &= \exp{({-\norm{\mathbf{a}_k - \mathbf{a}_{k-1}}}^2)}, \\
    r_{\square} &=  \exp{({-\norm{\square - \square^d}}^2)}
\end{align}
where $\square$ represents the measured value and $\square^d$ is the desired value. For the velocity rewards $r_{v_x}, r_{v_y}$, the target value is the desired walking speed sampled at the beginning of each episode. For the swing-foot rewards $r_{sw_x},r_{sw_y},r_{sw_z}$ the target values are the 3D foothold position, where the $(x,y)$ target coordinates are heuristically estimated from the desired walking speed, \revisedgc{e.g., $p_x=v_x^d/T$, $p_y=v_y^d/T + p_{y_{\mathrm{off}}}$ with $p_{y_\mathrm{off}}=\pm0.1$ being an offset to avoid feet collision in the lateral plane}, and the $z$ target coordinate is the terrain height corresponding to the $(x,y)$ coordinates. \revisedgc{The reward $r_{sw_h}$ encourages the policy to achieve sufficient swing-foot clearance, which is the maximum height the swing foot reaches during the swing phase. The reference value for swing-foot clearance is 5 cm above the terrain height at the swing foot's $(x,y)$ coordinates. This reward also helps to avoid unnecessarily over-lifting the swing foot when finding an obstacle. Finally, the reward $r_{sw_f}$ penalizes any contact force on the swing foot, which is used to encourage the policy to avoid early contact with the edges of the terrain during the swing phase. We denote that all the quantities used in the rewards are expressed in the stance-foot frame, which is a common approach in bipedal locomotion.} The weights for the reward terms are chosen as
 $\mathbf{w}^T = [0.2, 0.1, 0.075, 0.075, 0.15, 0.2, 0.15, 0.1]$. 

\subsection{RL training setup}
We use the Proximal Policy Optimization algorithm~\cite{schulman2017proximal} with input normalization, fixed covariance, and parallel experience collection to train the RL policy. 
The neural network selected for the RL policy is an MLP with two hidden layers, each with 128 units and $tanh$ activation function. We use a batch size of 64 for the PPO algorithm, a discount factor of 0.95, and 56000 samples with six epochs of policy update per algorithm iteration.
For each training episode, a terrain type is randomly selected from a set of different terrains: hills, slopes, random stairs, squared steps, and stairs up. These terrains are randomly generated from a diverse set of parameters, such as slope degree, number of stairs, stairs dimensions (width, height, depth), and size of squares, among others. Moreover, the initial state of the robot is drawn from a normal distribution about an initial pose corresponding to the robot standing in the double support phase. The same terrain parameter and initial state randomization are used during training, evaluation, and testing of the policies.

One iteration step of the policy corresponds to the interaction of the learning agent with the environment. The RL policy takes the reduced order state $\mathbf{s}$ and computes an action $\mathbf{a}$ converted into desired task-space trajectories at the time $t_k$. The reference trajectories are then sent to the task-space controller, which sends torque commands to the robot. This workflow is depicted in \figref{fig:overall_framework}. The feedback control loop runs at $1$ kHz, while the high-level planner policy runs at $33$ Hz. The maximum length of each episode is $300$ iteration steps, corresponding to $9$ seconds of simulated time.
An episode will be terminated early if the torso pitch and roll angles exceed $1$ rad or if the height of the robot's base relative to the stance foot is less than $0.4$ m.

\begin{figure}[t]
    \vspace{1mm}
    \begin{center} 
    \includegraphics[trim={0mm 0mm 0mm 0mm},clip,width=1.0\linewidth]{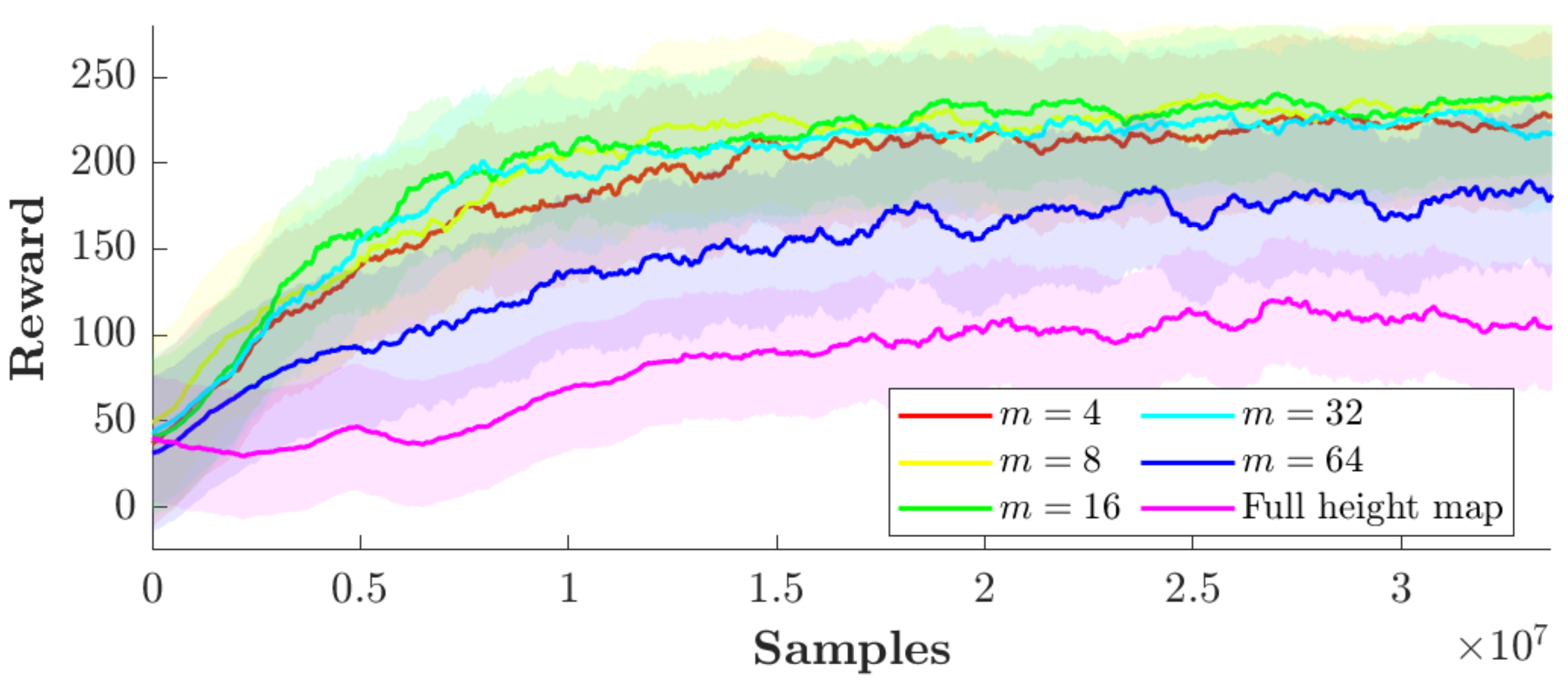}
    \end{center} 
    \caption{Reward convergence for different values of the latent variable dimension.
    Policies using an appropriately-sized latent, e.g., $m=16$, learn substantially faster and reach higher final rewards than those using an oversized latent, e.g., $m=64$ or the raw $800-dimensional$ height map.
    }
    \label{fig:learning_convergence}
\end{figure} 

\section{Simulation Results} \label{sec:sim_results}


\subsection{Learning convergence}
We demonstrate the effectiveness of the latent representation of the terrain by conducting an ablation study of the effect of this latent terrain representation on the efficiency and effectiveness of the learning process. In~\figref{fig:learning_convergence}, we present the evolution of the reward during the RL training for different values of the latent dimension $m$. Notably, the reward curves with better learning efficiency (fewer epochs to converge) correspond to $m \leq 32$, while the reward curves corresponding to $m\geq64$ show slower convergence to a lower value.
In addition, we replace the latent representation of the terrain with the complete local height map matrix, which results in significantly decreased sample efficiency and policy performance, as shown in~\figref{fig:learning_convergence}. 

We do not claim that using the full height map to train RL policies for locomotion successfully is unfeasible, but that an accurate selection of the dimension of the latent representation significantly increases the sample efficiency of the learning process, as demonstrated in ~\figref{fig:learning_convergence}.

Comparing our proposed framework with other baselines is not straightforward for two main reasons. First, to the best of our knowledge, our work is the first terrain-aware learning-based locomotion implemented for the robot Digit.

\revisedgc{Although the work in \cite{vanmarum2023learning, duan2023learning, gadde2025learning} shows similar approaches for learning-based perceptive locomotion, with the robot Cassie, these approaches are based on end-to-end and student-teacher RL frameworks that do not focus on sample efficiency. In these works, either the full local height map~\cite{duan2023learning}, samples of the terrain height~\cite{vanmarum2023learning}, or a distilled representation from several depth images~\cite{gadde2025learning} are used along with the full-order state of the robot as the inputs of the RL policy, and the output is the desired motor position.} While this approach is straightforward, it also significantly increases the complexity of the learning problem, requiring a higher number of parameters of the network, and a substantially higher number of samples to train a policy successfully. Table \ref{tab:comparison_frameworks} shows this comparison, exhibiting the advantages of our proposed approach with at least 2x increase in sample efficiency. 

\begin{table*}[ht]
\centering
\begin{tabular}{@{}lccccc@{}}
\toprule
\textbf{Method}  & \textbf{\# Samples}  & \textbf{Architecture} & \textbf{Perception input}  & \textbf{Policy output}     & \textbf{Pre-trained} \\ \midrule
\textbf{~\cite{vanmarum2023learning}}     & $60 \times 10^6$    & End-to-end    & Full height map           & Joint positions      & No                           \\
\textbf{~\cite{duan2023learning}}     & $60 \times 10^6$    & End-to-end    & Sampled terrain height    & Joint positions      & Yes                          \\
\textbf{~\cite{gadde2025learning}}     & $60 \times 1.4^9$    & Teacher-Student    & Sampled terrain height    & Joint positions      & Yes                          \\

\textbf{~Ours}    & $30 \times 10^6$     & Hierarchical  & Latent representation     & Task-space commands  & No                          \\ \bottomrule
\end{tabular}
\caption{\revisedgc{Comparison with other RL-based approaches for perceptive locomotion.}}
\label{tab:comparison_frameworks}
\end{table*}

\revisedgc{Second, the approach in \cite{duan2023learning, gadde2025learning} requires that the RL policy is already pretrained for blind locomotion, and their perceptive modules require a complex network architecture, including additional LSTM, CNN, ResNet, and U-NET networks. However, we acknowledge that these works have demonstrated successful sim-to-real transfer on the Cassie robot.}

In contrast, we propose an efficient yet effective framework that mainly focuses on our policies' sample efficiency and lightweight nature. This allows training policies from scratch with fewer samples (without pretrained policies or precomputed reference trajectories), while providing insightful observations about the importance of the dimension of the latent variable used to represent the terrain features, which is not addressed in other perceptive locomotion work. \revisedgc{We have not included other perceptive locomotion works, e.g. \cite{long2025learning, he2025attention, zhang2025distillationppo}, in this comparison as they do not specify information about the number of samples required for the RL training to converge.}

\subsection{Grid map resolution} \label{subsec:gridmapresolution}
\begin{figure}[t]
    \begin{center} 
    \includegraphics[width=1.0\columnwidth]{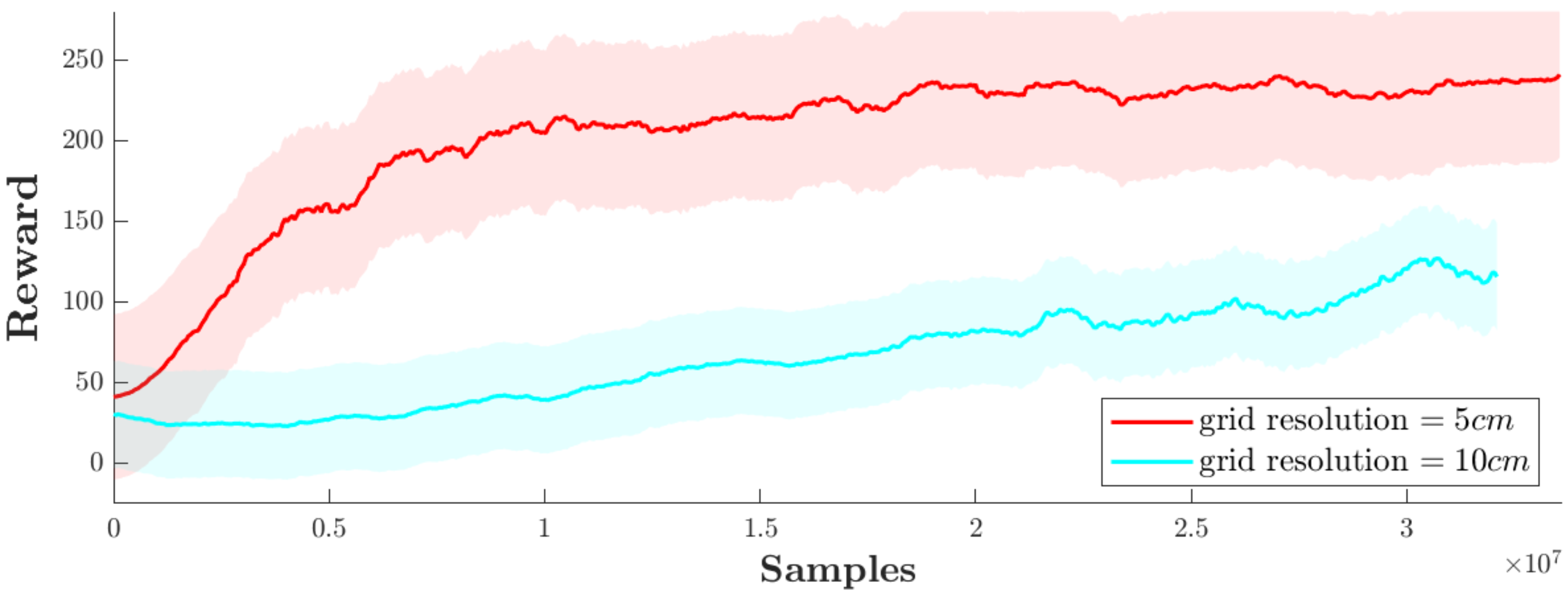}
    \end{center} 
    \caption{Comparison of depth learning performance with different resolutions of the height map grid, demonstrating that higher values of the terrain grid result in a degraded performance of the policy. }
    \label{fig:comparison_heigh_map_resolution}
\end{figure}

As mentioned in Section \ref{subsec:datacollection}, our intuition for the choice of grid size in the height map was to use a resolution as fine as the robot's foot width, which is about 5 cm. A grid of 5 cm is fine enough to capture the features of different types of terrain, while values higher than this, e.g., 10 cm, could be too coarse to capture important features like edges or borders of stairs and irregular terrains. This intuition also aligns with relevant work in the literature. For example, \cite{hoeller2023anymal} uses a variable resolution point cloud, where coarse resolution voxels (12.5 cm) are used to map the further scene of the robot. In comparison, high-resolution voxels (6.25 cm) capture the environment close to the robot. In \cite{vanmarum2023learning}, the terrain height is sampled using a pattern of 318 points adaptively spaced circularly around the foot position, where the samples close to the robot have a resolution of about 5 cm. Finally, in \cite{duan2023learning}, the authors also select a resolution of 5 cm for the height map grid, which is used as an input to an RL policy already pretrained for blind locomotion for the Cassie robot. Figure \ref{fig:comparison_heigh_map_resolution} shows that using a resolution of 5 cm results in significantly better rewards with fewer samples than 10 cm resolution. 

\subsection{Latent space reconstruction from depth images} \label{subsec:distillation}
The latent variable introduced in \eqref{eq:reparameterization} enables the policy to capture an efficient and effective representation of the terrain. However, deploying this approach on a physical robot poses a significant challenge: obtaining an accurate local height map can be computationally expensive, highly noise-sensitive, and often requires a costly sensor suite.

\revisedgc{To address the challenges in sim-to-real transfer for bipedal locomotion, we implement a latent space distillation framework that directly constructs the latent representation from raw depth sensor inputs. This approach circumvents the conventional dependency on local height maps, streamlining perception while enhancing robustness to real-world sensor noise.
Although several frameworks use the history of the depth images combined with the robot state in the distillation stage to reconstruct the entire terrain height map \cite{duan2023learning, hoeller2023anymal}, these approach results in complex network architectures and additional computational burden in training and inference of the perception module. In our work, we leverage synchronized feeds from two Intel D435i depth cameras mounted on the base of the robot's torso and pelvis to accurately recover the corresponding local height map from only one frame of the two combined depth images, ensuring sufficient coverage of the robot's surrounding terrain with minimal computational burden. A CNN-VAE is trained to align its latent space with that derived from previously trained local height map latent space. The training objective combines the VAE loss \eqref{eq:VAE_loss} for distributional regularization, an MSE loss between latent vectors from height-map processing (teacher) and from raw depth images (student).}
This additional distillation process is shown at the bottom in \figref{fig:overall_framework}. 

\begin{figure}[t]
    \begin{center} 
    \includegraphics[width=1\linewidth]{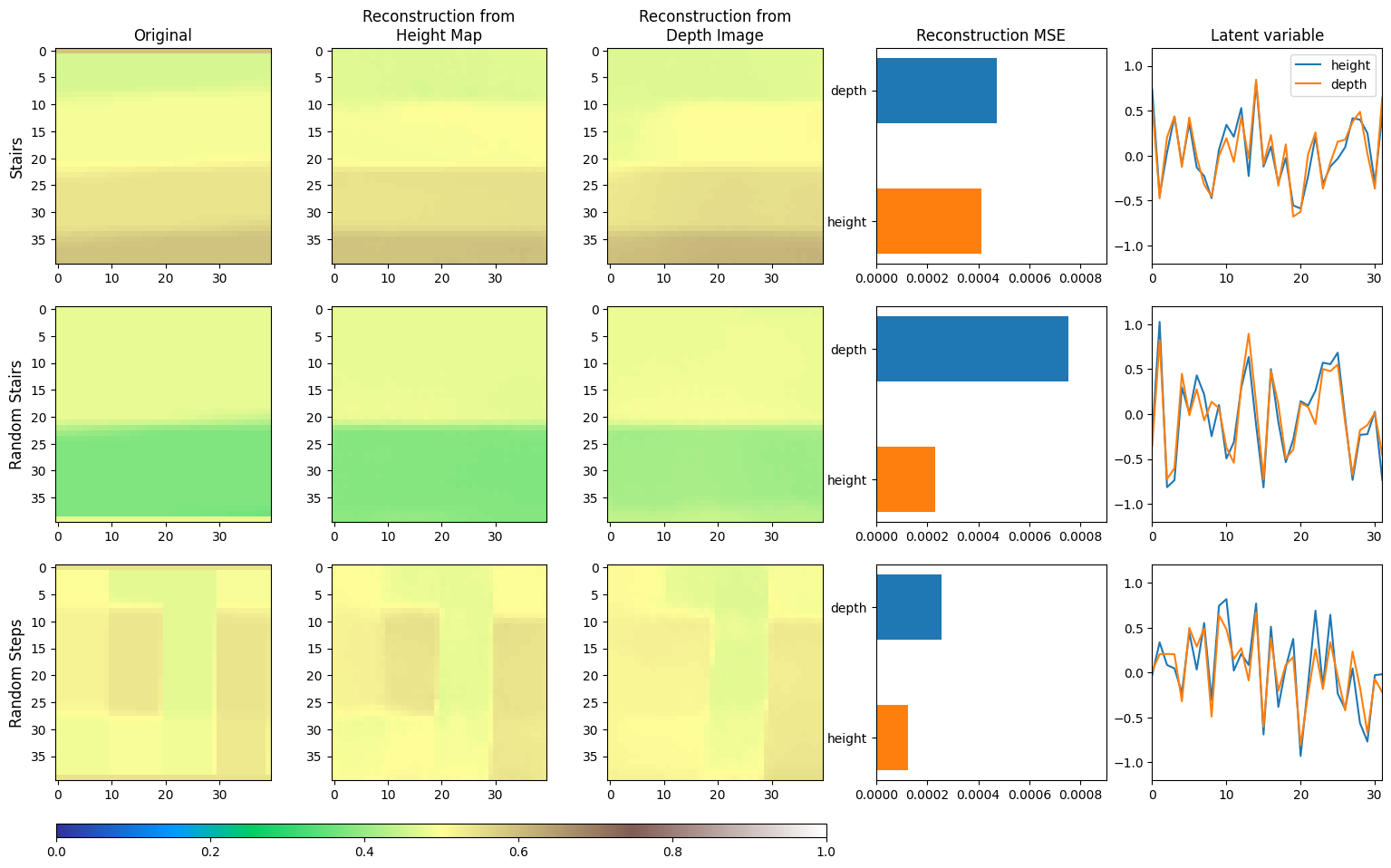}
    \end{center} 
    \caption{Reconstruction of local height maps from latent encodings learned from two different exteroceptive sources. Each column shows: (1) the ground-truth height map for a sample terrain; (2) the reconstruction from the latent vector produced by the height map encoder; (3) the reconstruction from the latent vector produced by the depth image encoder; (4) the error between the two reconstructions, and (5) the latent variable comparison. The reconstruction error is very low, demonstrating that the depth-image encoder successfully captures a similar latent representation to the height-map-based encoder.} 
    \label{fig:reconstruction_height_maps_terrains}
\end{figure}

\figref{fig:reconstruction_height_maps_terrains} analyzes the accuracy of the reconstructions of the local height map using the latent variable obtained from: i) the original local height map and ii)the depth camera images. The first column in \figref{fig:reconstruction_height_maps_terrains} shows the ground truth of height map samples of different terrains (Stairs, Random Stairs, and Squared Steps) obtained from simulation. The second and third columns show the reconstruction of the local height maps from the latent variable encoded directly from the local height matrix and the latent variable encoded from the depth images, respectively. The latent variable from the local height map and the depth images is depicted in the last column, where we show the effectiveness of the distillation process in learning the same latent representation from two different sources.
Finally, while the visual comparison of the images in the second and third columns clearly shows a good reconstruction of the terrain's height map, we quantitatively capture the accuracy of the reconstruction from the latent variable by showing the MSE error between the two height-map matrices in the fourth column of \figref{fig:reconstruction_height_maps_terrains}. 

\begin{figure*}[t]
    \begin{center} 
    \includegraphics[trim={1mm 1mm 1mm 1mm},clip,width=2\columnwidth]{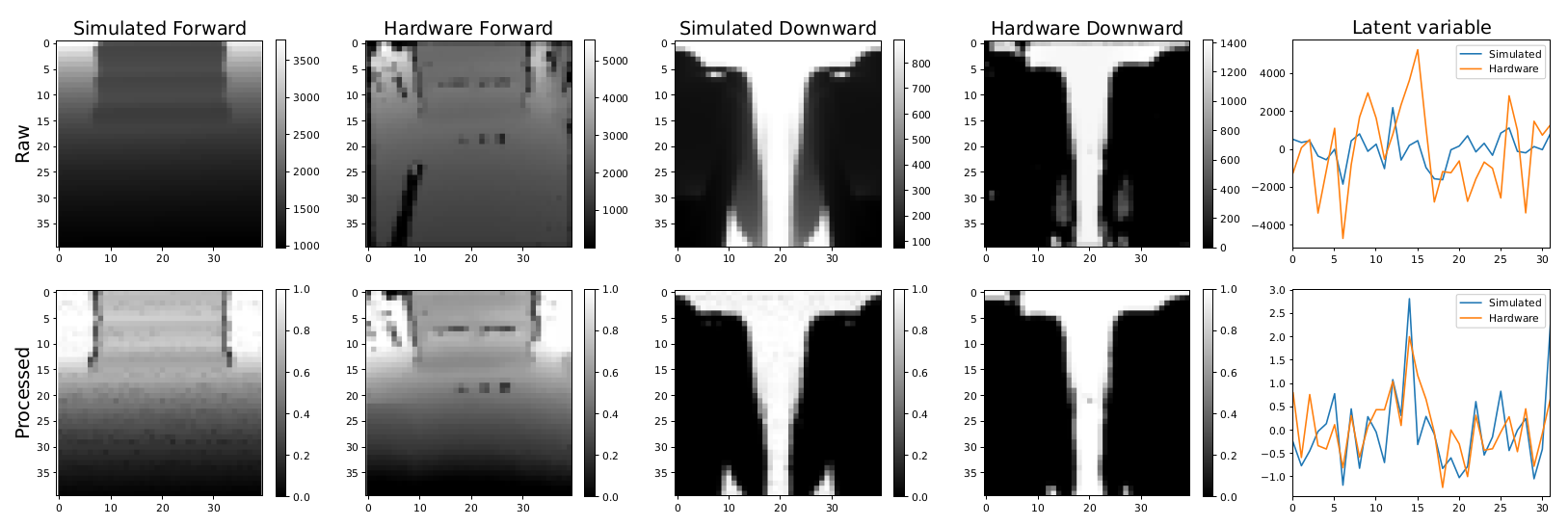}
    \end{center} 
    \caption{\revisedgc{Comparison of depth images from the hardware and the simulation before and after processing, e.g, clipping, cropping, filtering.  The top row presents the raw data obtained from each source, whereas the bottom row depicts simulated and hardware post-processed depth images. The processed simulated depth image closely resembles the real sensor image, and their resulting latent vectors are also very similar. This shows that with noise and occlusion handling, our simulation model produces depth data that the encoder perceives as real data, supporting the potential for sim-to-real transfer.}} 
    \label{fig:comparison_depth_images}
\end{figure*} 


\revisedgc{Moreover, to reduce the sim-to-real gap caused by the difference between the depth images in simulation and hardware, we process the simulated depth images using post-processing techniques inspired by \cite{gadde2025learning, rudin2025parkour} that have demonstrated successful sim-to-real transfer of perceptive-locomotion policies. In particular, we i) crop the image to remove the blind spots cause by stereo-matching at small distances, ii) add Gaussian blurring and characteristic depth shadowing around edges using Canny edge detectors, iii) clipped max depth to 2 m, iv) inpainting for hole filling with edge coherence, v) edge pixel dropout simulating occlusion artifacts, and vi) masking of big occlusions caused by the legs in the downward camera. To demonstrate the effectiveness of these techniques, \figref{fig:comparison_depth_images} shows a comparison between simulated and real camera feedback from the D435i cameras on the Digit robot, highlighting the resemblance of the simulated environment.}


\begin{table*}[ht]
\centering
\begin{tabular}{@{}lccccc@{}}
\toprule
\textbf{\revisedgc{Terrain Type}} & \textbf{\revisedgc{Parameter}} & \textbf{\revisedgc{Distribution}} & \textbf{\revisedgc{Params (Mean, Std)}} & \textbf{\revisedgc{Limits (Min, Max)}} & \textbf{\revisedgc{Ext Limits (Min, Max)}} \\ \midrule
\revisedgc{Stairs} & \revisedgc{Step length [m]} & \revisedgc{Normal} & \revisedgc{(0.4, 0.1)} & \revisedgc{(0.3, 0.5)} & \revisedgc{(0.25, 0.55)} \\
                   & \revisedgc{Step height [m]} & \revisedgc{Normal} & \revisedgc{(0.15, 0.1)} & \revisedgc{(0.05, 0.25)} & \revisedgc{(0.025, 0.275)} \\
\revisedgc{Hills}  & \revisedgc{Max amplitude}   & \revisedgc{Uniform} & \revisedgc{N/A} & \revisedgc{(0.3, 0.4)} & \revisedgc{(0.26, 0.44)} \\
                   & \revisedgc{Octaves}         & \revisedgc{Normal} & \revisedgc{(10.0, 2.0)} & \revisedgc{(5, 15)} & \revisedgc{(3.5, 16.5)} \\
\revisedgc{Slopes} & \revisedgc{Slope plane about axis x} & \revisedgc{Normal} & \revisedgc{(0.0, 0.1)} & \revisedgc{(-0.2, 0.2)} & \revisedgc{(-0.22, 0.22)} \\
                   & \revisedgc{Slope plane about axis y} & \revisedgc{Normal} & \revisedgc{(0.0, 0.1)} & \revisedgc{(-0.2, 0.2)} & \revisedgc{(-0.22, 0.22)} \\
\revisedgc{Square Steps} & \revisedgc{Max step size [m]}   & \revisedgc{Uniform} & \revisedgc{N/A} & \revisedgc{(0.3, 0.5)} & \revisedgc{(0.25, 0.55)} \\
                   & \revisedgc{Max step height [m]} & \revisedgc{Uniform} & \revisedgc{N/A} & \revisedgc{(0.15, 0.25)} & \revisedgc{(0.125, 0.275)} \\ \bottomrule
\end{tabular}
\caption{\revisedgc{Parameters sampled from distributions for different terrains.}}
\label{tab:terrain_parameters}
\end{table*}

\subsection{Policy performance}
We denote that the RL policy learns to walk from scratch and that one trained policy can successfully navigate on various terrains, as shown in \figref{fig:tileplots_terrain}.
When walking over irregular terrain, the policy adapts the foot landing location by taking shorter or longer steps to avoid collisions with the edges of the terrain. Similarly, the policy adapts the foot location to compensate for the heavy robot's inertia to prevent falling, i.e., when stepping up or down the stairs. These adaptive strategies naturally emerged during the training from effectively integrating the terrain features into the RL policy and the combination of rewards. \revisedgc{We denote all results presented using a fixed step duration of $0.4$ seconds. The policy does not change the step timing in response to terrain but adjusts foot placement. While adding a variable stepping frequency could be an interesting extension, the primary focus of this work is on the impact of the latent representation of the terrain.} More details about the policy performance can be seen in the accompanying video: \colorlink{\href{https://youtu.be/tJVfQK2XcQs}{https://youtu.be/tJVfQK2XcQs}}.

\begin{figure}[t]
    \vspace{1mm}
    \begin{center} 
    \includegraphics[width=0.95\linewidth]{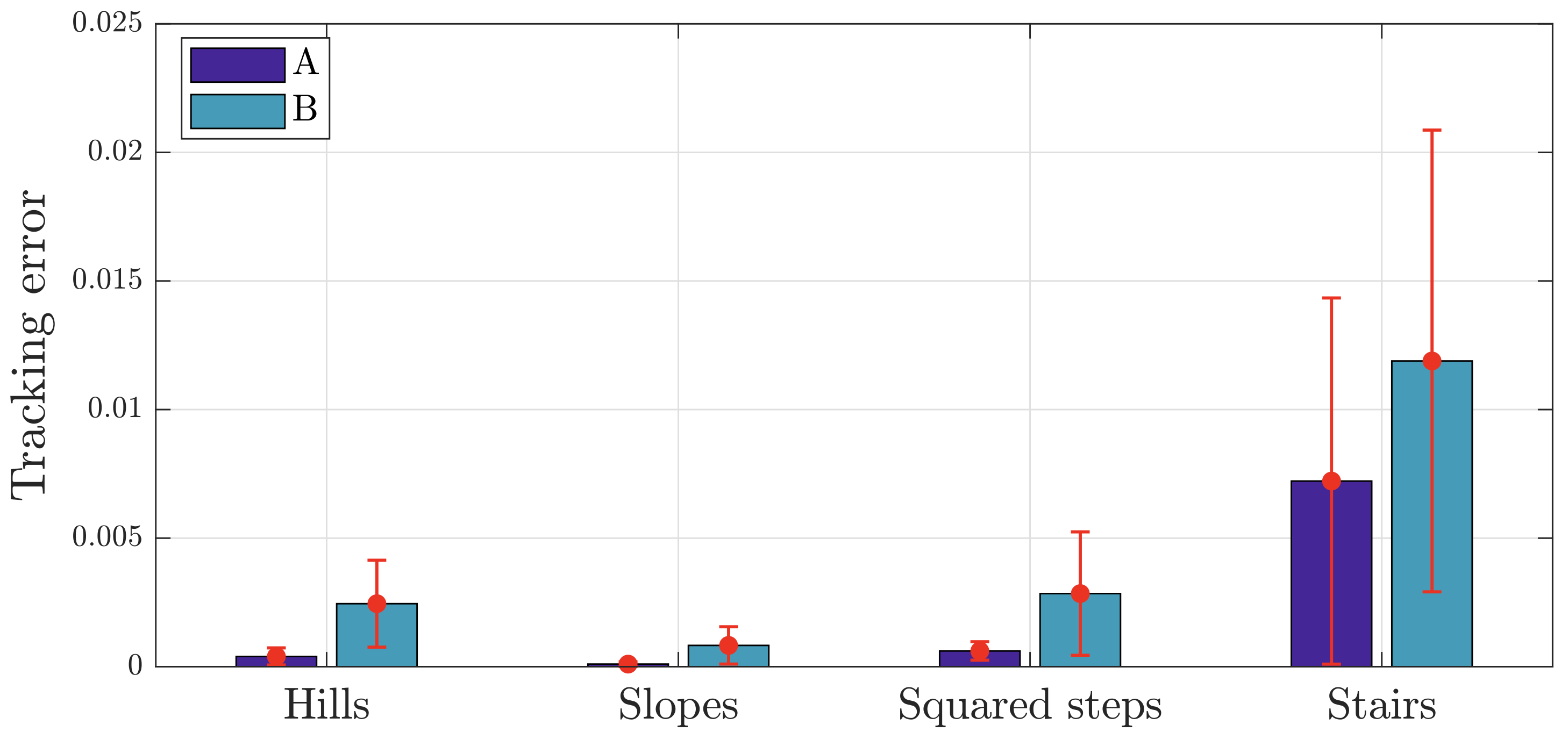}
    \end{center} 
    \caption{Average MSE for velocity tracking over 20 runs for height map policy (A) and depth image policy (B).}\label{fig:speed_tracking_different_terrains}
\end{figure}

Table \ref{tab:terrain_parameters} shows the range of terrain parameters used during data collection, policy training, and evaluation. These parameters were consistent across all the experiments to ensure a fair comparison and to avoid domain gaps between training and testing. 
\revisedgc{Although the policy navigates successfully on different terrains, there is a trade-off between robustness and tracking the commanded walking speed as the complexity of the environment causes a higher tracking error in some of the terrains. We denote that the velocity tracking reward is a soft constraint within the RL formulation; therefore, perfect tracking is not expected, especially when it could come at a higher cost for other important rewards, e.g., avoiding the robot from falling, which would result in an early episode termination. In other words, the policy sacrifices velocity tracking performance to guarantee the robustness of the walking gait.}
This is particularly evident in \figref{fig:speed_tracking_different_terrains}, where we show the tracking error between the average walking speed $\hat{v}_x$ and the desired walking speed $v_x^d=0.5 m/s$ over 20 runs of the same policy for four different terrains. Despite irregularities in the terrain, the policy adapts its behavior to keep close track of the target speed, except in cases where the terrain conditions are too challenging, e.g., steep stairs, forcing the policy to deviate from the desired walking speed to avoid falling. \revisedgc{Even with terrain awareness, the robot must exert greater corrective control on more difficult terrains, which leads to higher tracking error. The latent representation guides foot placement to appropriate locations. However, once a foot makes contact on an irregular surface, disturbances (like small slips or tilts) can still occur and must be corrected by the controller, resulting in deviations. Moreover, on more challenging terrains, the robot’s dynamics are more perturbed – for example, when a foot lands on a high step, the robot’s body might experience a jolt or require more corrective effort, leading to larger tracking errors in velocities.
}

\subsection{Robustness and comparison} \label{subsec:robustness}

\begin{figure}[b]
    \begin{center} 
    \includegraphics[width=0.95\linewidth]{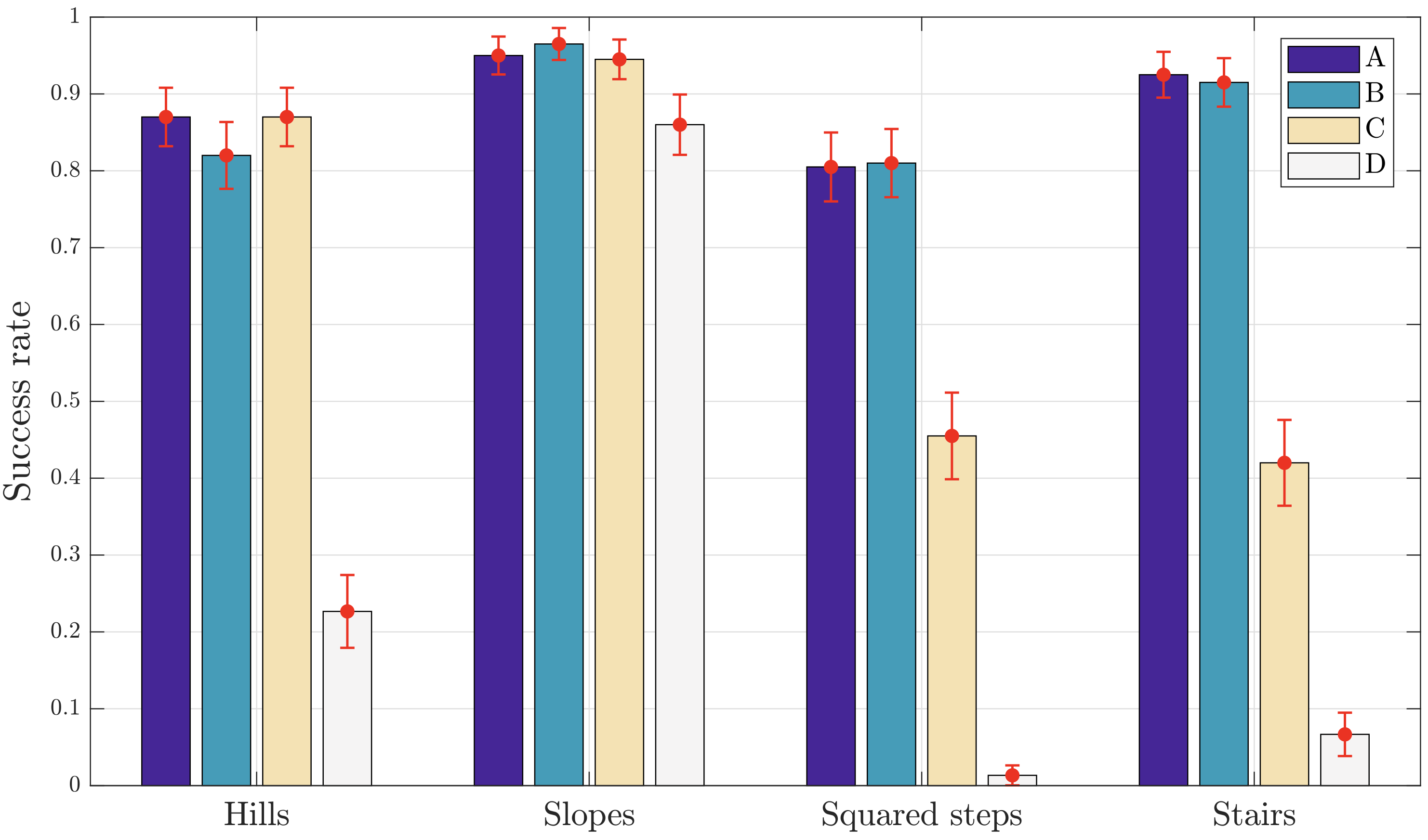}
    \end{center} 
    \caption{Robustness of the policy to different terrains collected over more than 200 experiments per terrain with a confidence interval $\geq 95 \%$.} 
\label{fig:success_rate}
\end{figure}

\revisedgc{To quantitatively assess the policy's robustness, we conduct a Monte Carlo evaluation. We test the policy's performance across diverse terrains for at least two hundred experiments per terrain type, where each terrain instance is randomly generated by sampling its parameters from the distributions shown in Table II.} The success rate for each terrain type is shown in \figref{fig:success_rate} with a confidence interval $\geq 95\%$. A successful trial consists of the robot walking without falling for 9 seconds. This consistency of success across a spectrum of terrains highlights the capability of the policy to navigate effectively and adapt without terrain-specific tuning. These results also demonstrate the policy's reliability in challenging terrains, an essential quality for humanoid robots to promote real-world deployment. 

Furthermore, to demonstrate the actual contribution of the latent representation of the height map, we compare our proposed approaches (policy A and B) with two baselines that share the same RL policy structure but use different inputs for the terrain representation. We denote these two baselines as policy C and policy D. 

Policy C corresponds to the case of blind locomotion, where the policy does not have a meaningful representation of the upcoming terrain. By keeping the input of the local height map to a fixed value corresponding to flat terrain, the policy "thinks" that it is walking on flat ground. 
The policy is robust enough to handle terrains with hills and slopes, which is consistent with several works on blind bipedal locomotion, where it has been shown that the potential of RL policies to navigate on these types of terrains without using exteroceptive feedback. However, its success rate drops significantly for terrains with random square steps and stairs, resulting in the robot immediately falling after tripping over the edges of the steps in the terrain. 

Policy D corresponds to the case where the full terrain height map $\mathbf{x} \in \mathbb{R}^{20 \times 40}$ is included in the policy state. While this approach has been successfully applied in other end-to-end RL frameworks for bipedal locomotion~\cite{vanmarum2023learning,duan2023learning}, it is incompatible with our proposed compact and sample-efficient framework. We hypothesize that the lack of structure in the raw terrain height map data results in a bottleneck for learning effective actions. This effect is observed in~\figref{fig:learning_convergence}, where the reward curve for the policy with the full height map converges significantly slower to a smaller value than the policies that use the latent representation of the terrain height map. To alleviate this effect,~\cite{duan2023learning} builds upon a pre-trained RL policy according to~\cite{siekmann2021sim} and uses the complete height map along with the full-order robot's state to learn compensations added to the base RL policy. As shown in \figref{fig:success_rate}, policy D is the worst performer, even under-performing blind locomotion (policy C).

On the other hand, policy A, which corresponds to our approach of training the policy using the reduced latent representation of the local height map, and policy B, which corresponds to the approach of generating the latent representation of the local height map using depth cameras, perform consistently across all terrain types with a success rate greater than $80\%$. This underlines the robustness of reduced-order latent representations and their capabilities to generalize well across different terrains without fine-tuning and to generalize to different perceptive sources, i.e., depth cameras, through a simple distillation process. 

\subsection{\revisedgc{Generalization to out of distribution parameters and sim-to-sim transfer}}

\revisedgc{To evaluate the generalization capability of the learned policy beyond the training distribution, we conducted additional tests using an extended range of terrain parameters not seen during training. Specifically, we increased the upper bounds of the terrain generation parameters by $20\%$ across all terrain types, as detailed in the last column of Table~\ref{tab:terrain_parameters}. This controlled extrapolation aims to assess the robustness of the policy under terrain conditions that exceed the complexity of the training set.}

\begin{figure}[t]
    \begin{center} 
    \includegraphics[width=1.0\linewidth]{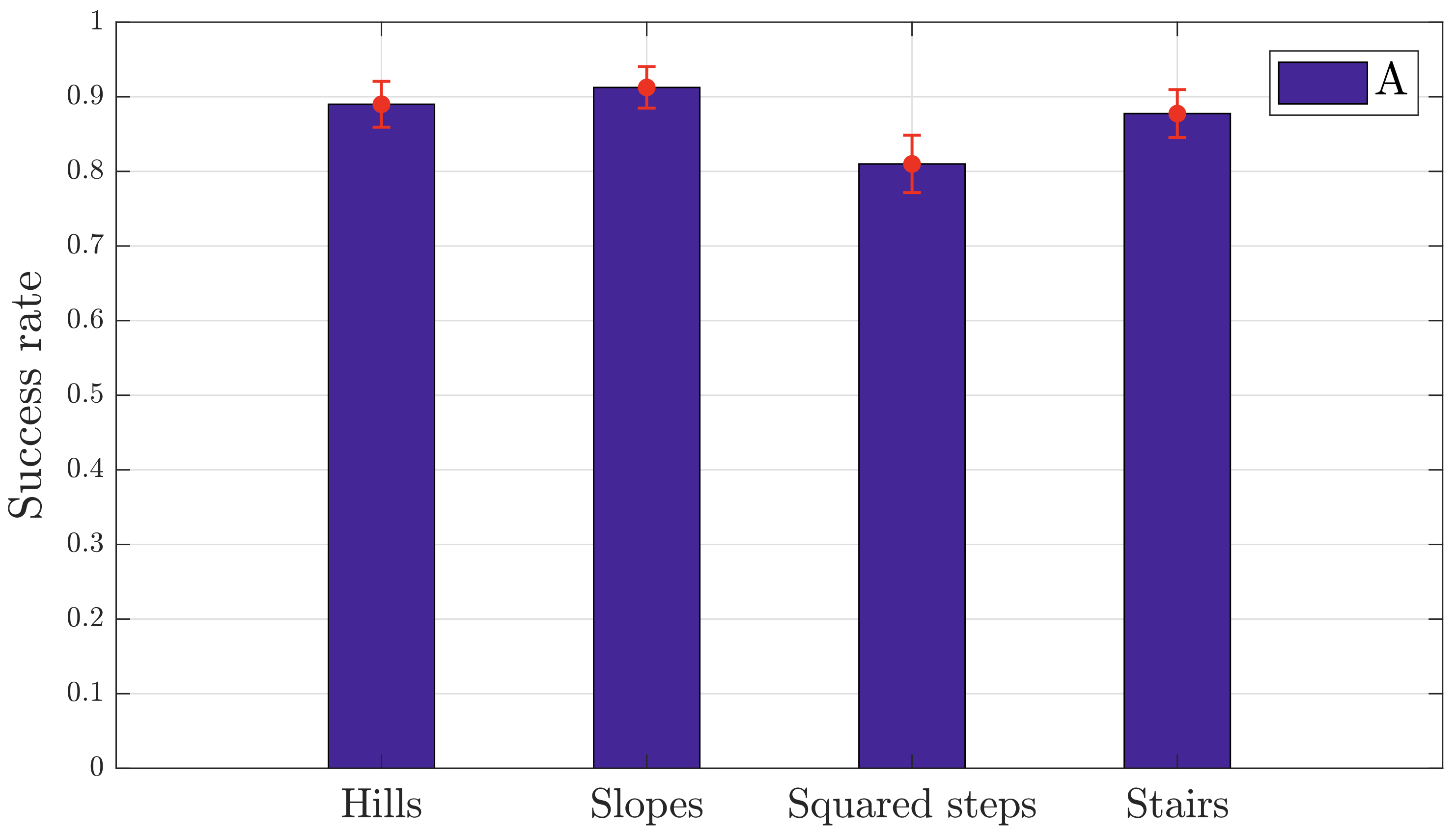}
    \end{center} 
    \caption{\revisedgc{Success rates with out-of-distribution terrains with an extended range ($20\%$) of key terrain parameters concerning the original training range. The extended limits are presented in Table \ref{tab:terrain_parameters}. The policy maintains high success on most terrains (e.g., hills, squared steps) despite the increased difficulty, highlighting the robustness of the learned perceptual latent space and the policy’s adaptability.}}
\label{fig:success_rate_extended_params}
\end{figure}

\revisedgc{As shown in Fig.~\ref{fig:success_rate_extended_params}, the policy maintains high success rates on terrains such as \textit{hills}, and \textit{squared steps}, despite the increased difficulty. We denote that the CNN-VAE module used for terrain encoding was not retrained on these extended terrains. The strong performance, therefore, highlights both (i) the robustness of the learned perceptual latent space in encoding previously unseen terrains, and (ii) the adaptability of the policy to respond effectively to out-of-distribution scenarios. These results demonstrate that the policy has acquired transferable perceptual-motor representations capable of generalizing beyond the training distribution for various terrain types.}

\revisedgc{In contrast, performance on the \textit{slopes} and \textit{stairs} terrains slightly degrades under the extended parameter settings. We attribute this to the structured and discontinuous nature of stair environments, which become significantly more challenging with increased step height and gap width and the complex interaction between the flat landing foot and the step slopes. These changes represent a challenge for the kinematic and dynamic limitations of the Digit robot, reducing the available foothold margin and increasing the likelihood of unstable contact or foot scuffing during swing and landing. While out-of-distribution generalization is not the primary focus of this work, we recognize the value of incorporating more complex mechanisms—such as the attention-based models applied in ~\cite{he2025attention}—to better handle such terrain complexities or online adaptive strategies for foot orientation. We consider this an important direction for future work.}

Finally, we successfully tested policy A in the Agility Robotics Simulator, a highly realistic environment for the Digit robot, including features such as real-time simulation, communication delays, actuator delays, and the exact state estimation used in the hardware. It also shares the same API as the hardware, meaning that the same code used in the simulation can be deployed on the hardware with a high probability of success. The effectiveness of the AR simulator as a good sim-to-real evaluation tool has been demonstrated in several works using Digit, where policies tested in the AR simulator have been successfully transferred to hardware \cite{castillo2021robust, paredes2024safe, shamsah2023integrated}.

\begin{figure}[t]
    \begin{center} 
    \vspace{4mm}    
    \scalebox{-1}[1]{\includegraphics[width=1.0\columnwidth]{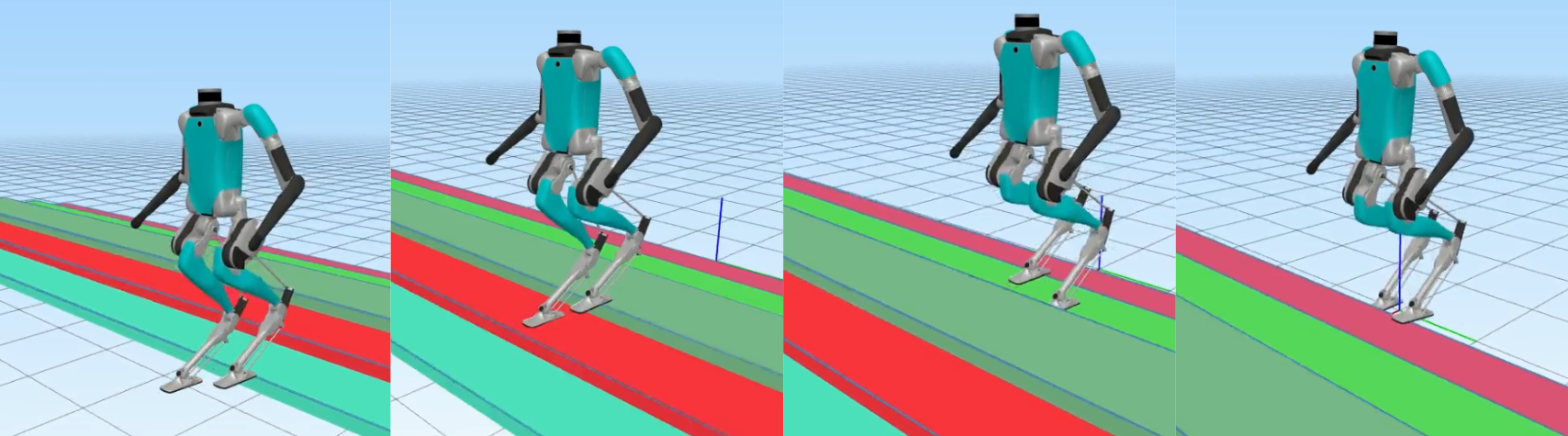}}
    \end{center} 
    \caption{A sequence of images of Digit walking up and down a staircase in the Agility Robotics high-fidelity simulator. Using our terrain-aware policy, the robot successfully climbs and descends the stairs without falling.} 
    \label{fig:digit_stairs_arsim}
\end{figure}

In \figref{fig:digit_stairs_arsim}, we show a tile plot of the robot walking up and down stairs in the AR simulator. Since the AR simulation does not provide the depth camera feedback, the policy shown in \figref{fig:digit_stairs_arsim} corresponds to policy A in Section \ref{subsec:robustness}. The latent representation is encoded from the local terrain height map. The policy successfully leverages the efficient latent representation of the terrain to command the task space actions that allow the robot to lift its feet at the right time and place to successfully traverse the stairs without falling. A detailed sequence of the motion is also shown in the accompanying video.
In addition, to verify the importance of the latent representation of the terrain, we also test policy C in the AR simulator. As expected and consistent with the results in Section \ref{subsec:robustness} and \figref{fig:success_rate}, the blind policy falls when the foot hits the edge of the stairs. These results are also presented in the accompanying video.


\revisedgc{\subsection{Comparison with History-Aware Perception} \label{subsec:history_results}
Finally, we show the performance of the policies trained using the latent representation based on the reconstruction results of the new CNN-VAEs CNN-VAE training loss curves in Section \ref{subsec:history_method}, \figref{fig:CVAE_history_train_loss_compare}, which indicates successful convergence for most history-aware CNN-VAEs.}

\begin{figure}[t]
    \begin{center} 
    \includegraphics[trim={16mm 8mm 20mm 10mm},clip,width=1\columnwidth]{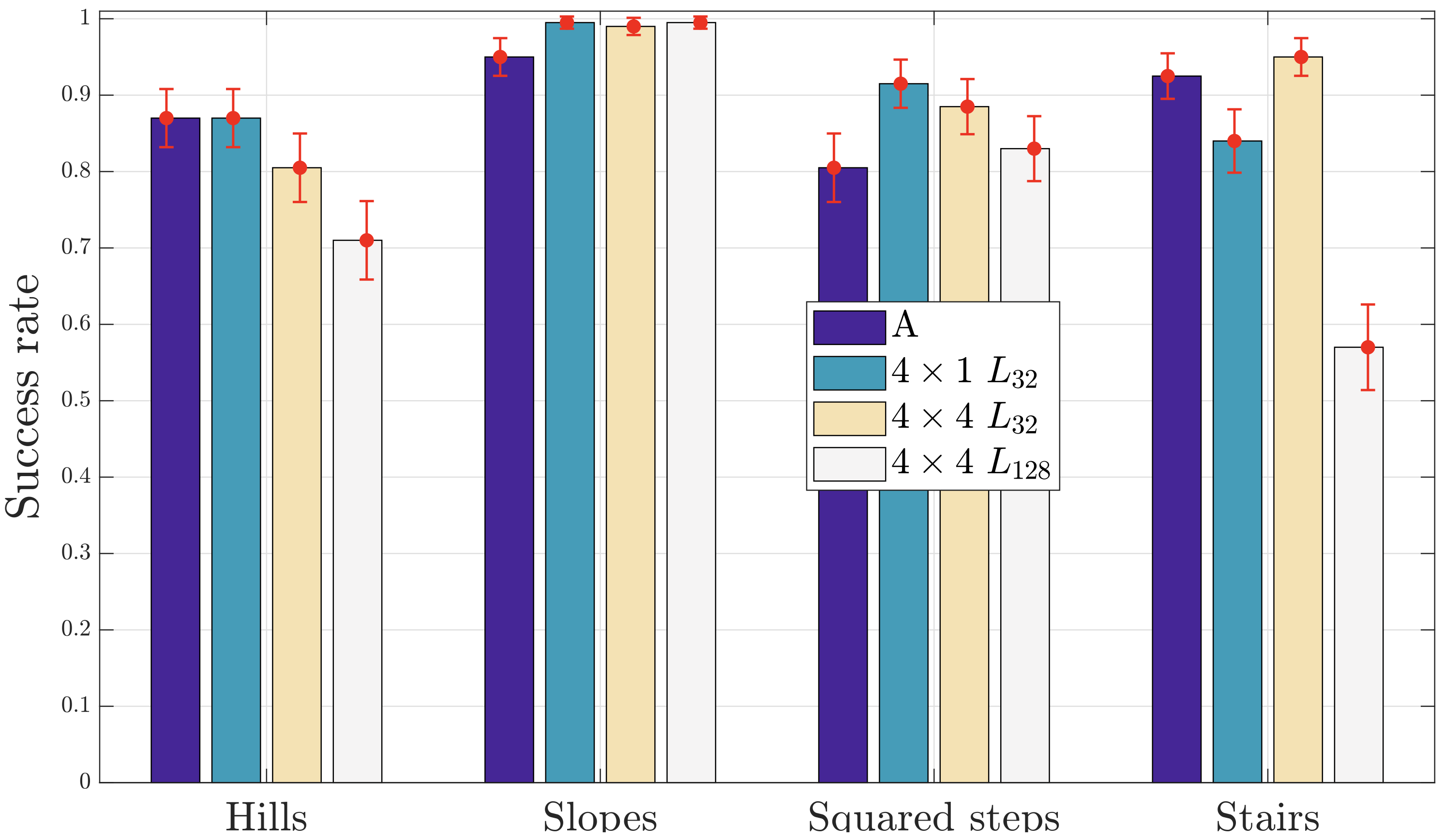}
    \end{center} 
    \caption{\revisedgc{Success rates on various terrains for policies trained with history-augmented terrain encoders, comparing different input-output structures and latent sizes. Including a short history of height maps can improve robustness on certain terrains (all history-based models exceed $95\%$ success on slopes). However, using a larger latent space, e.g., $L_{64}$ vs. $L_{32}$, does not guarantee better performance. In fact, the $4\times1 L_{32}$ model (4-frame history compressed into one latent of size 32) emerged as the best performer on most terrains.}} 
    \label{fig:success_rate_history}
\end{figure}

\revisedgc{Success rates for history-aware policies across terrains (\figref{fig:success_rate_history}) show variations in effectiveness depending on the terrain type. While all models excelled on slopes (success rates $> 95\%$), performance differences were more pronounced on complex features like hills and stairs.}

\revisedgc{Despite the superior reconstruction accuracy, larger latent spaces ($L_{64}$, $L_{128}$) did not always result in better locomotion policies. Despite its compressed representation, the $4 \times 4 \hspace{2mm} L_{32}$ model consistently performed well, indicating that dimensionality reduction preserves essential terrain features while filtering noise. Interestingly, the $4 \times 1 \hspace{2mm} L_{32}$ model, which averages temporal inputs, emerged as the best performer across terrains, except for stairs, where its success rate dropped to $57\%$. This counterintuitive result suggests that for most terrains, a temporally compressed representation capturing the average terrain ahead provides robustness for locomotion planning.}

\revisedgc{We hypothesize that the averaging effect in the $4 \times 1 L_{32}$ architecture acts as an implicit regularization mechanism, enhancing policy robustness by emphasizing persistent terrain features over transient details that can act as a predictor of the incoming terrain.  This benefits navigation on slopes, hills, and squared steps, where gradual transitions outweigh precise height details. For example, when walking on a slope with a fixed inclination, the history of the height maps could help the policy infer the slope of the terrain, so it could adjust its stepping with the assumption that the incoming terrain in one or two steps could share the same terrain features as the past terrain. However, for stair traversal, where step height and edge detection are critical, temporal compressing blurs essential features, significantly hindering performance.}


\section{Conclusion} 
\label{sec:conclusion}
We propose a framework for learning terrain-aware perceptive locomotion that integrates a latent representation of the local height map with a reduced-order representation of the robot's states to form an efficient state representation. By combining a learning-based high-level terrain-aware planner that formulates effective task-space actions with a low-level feedback tracking controller, we obtain a robust controller capable of traversing challenging terrains while preserving excellent speed-tracking performance. 

A central contribution of this work is the detailed analysis of the latent space dimension, with ablation studies providing empirical demonstrations that a larger dimension is not necessarily better at capturing meaningful terrain features. Our investigation into this principle of minimal sufficiency for perceptual information revealed that an optimally compressed latent representation is critical for sample efficiency and policy robustness. This principle was further validated through an analysis of history-aware perception.

We have established a clear and promising path toward real-world application by successfully distilling this compact representation from depth camera images with realistic sensor noise and validating our policy in the high-fidelity Agility Robotics simulator. Future work will focus on the direct hardware implementation of this framework on the Digit robot, building on the strong sim-to-real evidence presented. Further investigation will also explore adaptive mechanisms to handle more extreme out-of-distribution terrains, pushing the boundaries of agile and perceptive locomotion.

\bibliographystyle{IEEEtran}
\bibliography{references}

\end{document}